\documentclass[10pt,twocolumn,letterpaper]{article}

\usepackage{cvpr}              

\renewcommand{\paragraph}[1]{\vspace{.5em}\noindent\textbf{#1}}

\usepackage[accsupp]{axessibility}  
\definecolor{cvprblue}{rgb}{0.21,0.49,0.74}
\usepackage[pagebackref,breaklinks,colorlinks,allcolors=cvprblue]{hyperref}
\usepackage{booktabs}
\usepackage{makecell}
\usepackage{multirow}
\usepackage{pifont}      
\usepackage{graphicx}    
\usepackage[table]{xcolor}
\usepackage{footnote}
\newcolumntype{G}{>{\columncolor{gray!10}}c} 
\newcolumntype{H}{>{\columncolor{gray!25}}c} 
\newcolumntype{I}{>{\columncolor{gray!10}}c} 

\makeatletter
\newcommand{\thickhline}{%
 \noalign {\ifnum 0=`}\fi \hrule height 1pt
 \futurelet \reserved@a \@xhline
}
\makeatother

\newcommand{\cmark}{\ding{51}}  
\definecolor{correctgreen}{RGB}{50, 168, 82}
\definecolor{wrongred}{RGB}{226, 58, 58}
\usepackage{colortbl}
\definecolor{secondcolor}{RGB}{189,215,238}
\definecolor{firstcolor}{RGB}{255,153,153}
\newcommand{\firstcolor}[1]{\cellcolor[rgb]{1,.60,.60}{#1}}
\newcommand{\secondcolor}[1]{\cellcolor[rgb]{.741,.843,.933}{#1}}
\usepackage{makecell}

\def\paperID{15191} 
\def\confName{CVPR}
\def\confYear{2026}

\title{Multimodal Protein Language Models for Enzyme Kinetic Parameters:\\From Substrate Recognition to Conformational Adaptation}
\author{Fei Wang$^{1,2}$, Xinye Zheng$^{1}$, Kun Li$^3$, Yanyan Wei$^{1,4,}$\thanks{Corresponding authors}, Yuxin Liu$^{2}$, Ganpeng Hu$^{5}$,\\  Tong Bao$^{5}$, Jingwen Yang$^{5}$\\
\normalsize$^1$ School of Computer Science and Information Engineering, Hefei University of Technology \\
\normalsize$^2$ Institute of Artificial Intelligence, Hefei Comprehensive National Science Center \\
\normalsize$^3$ 
CVLab, College of Information Technology, 
United Arab Emirates University  \\
\normalsize$^4$ Intelligent Interconnected Systems Laboratory of Anhui Province \\
\normalsize$^5$ School of Food Biological Engineering, Hefei University of Technology \\
}

\begin{document}
\twocolumn[{
\maketitle
\begin{center}
\captionsetup{type=figure}
\vspace{-25pt}
\includegraphics[width=\textwidth]{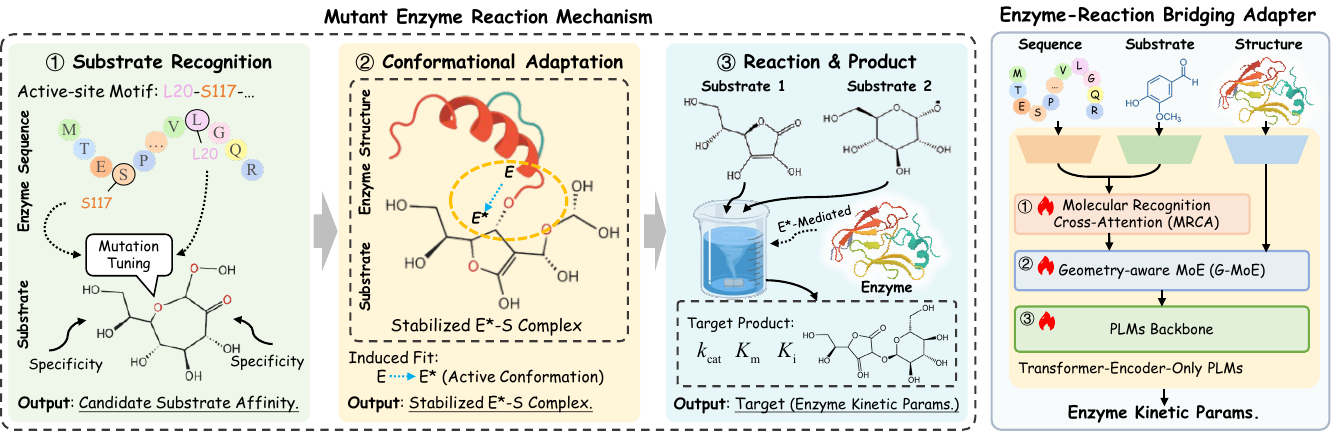}
\vspace{-2em}
\caption{
\textbf{Overview of the \textit{Mutant Enzyme Reaction Mechanism} and the proposed \textit{Enzyme-Reaction Bridging Adapter}.}
Catalysis proceeds through three stages: (1) Substrate recognition, tuning enzyme-substrate specificity; (2) Conformational adaptation, forming a stabilized $E^*$-$S$ complex; and (3) Reaction and product formation, yielding kinetic parameters ($k_\text{cat}$, $K_\text{m}$, $K_\text{i}$).
ERBA mirrors this process through Molecular Recognition Cross-Attention (MRCA) for sequence-substrate conditioning and Geometry-aware Mixture-of-Experts (G-MoE) for structural adaptation, enabling biologically grounded enzyme kinetics prediction.
}
\label{fig:fig1}
\end{center}
\vspace{-0.4em}
}]


\begin{abstract}
Predicting enzyme kinetic parameters quantifies how efficiently an enzyme catalyzes a specific substrate under defined biochemical conditions. Canonical parameters such as the turnover number ($k_\text{cat}$), Michaelis constant ($K_\text{m}$), and inhibition constant ($K_\text{i}$) depend jointly on the enzyme sequence, the substrate chemistry, and the conformational adaptation of the active site during binding. Many learning pipelines simplify this process to a static compatibility problem between the enzyme and substrate, fusing their representations through shallow operations and regressing a single value. Such formulations overlook the staged nature of catalysis, which involves both substrate recognition and conformational adaptation. In this regard, we reformulate kinetic prediction as a staged multimodal conditional modeling problem and introduce the Enzyme-Reaction Bridging Adapter (ERBA), which injects cross-modal information via fine-tuning into Protein Language Models (PLMs) while preserving their biochemical priors. ERBA performs conditioning in two stages: Molecular Recognition Cross-Attention (MRCA) first injects substrate information into the enzyme representation to capture specificity; Geometry-aware Mixture-of-Experts (G-MoE) then integrates active-site structure and routes samples to pocket-specialized experts to reflect induced fit. To maintain semantic fidelity, Enzyme-Substrate Distribution Alignment (ESDA) enforces distributional consistency within the PLM manifold in a reproducing kernel Hilbert space. Experiments across three kinetic endpoints and multiple PLM backbones, ERBA delivers consistent gains and stronger out-of-distribution performance compared with sequence-only and shallow-fusion baselines, offering a biologically grounded route to scalable kinetic prediction and a foundation for adding cofactors, mutations, and time-resolved structural cues.
\end{abstract}

\section{Introduction}
\label{sec:intro}
High-throughput protein design and synthetic biology~\cite{markin2021revealing, craven2018high, buller2023nature, chen2025glycolysis, kraut2003challenges, wolfenden2001depth, kaiser1984chemical} increasingly depend on \emph{in silico} estimates of the efficiency with which an enzyme converts a given substrate under realistic biochemical conditions. Turning this promise into practice requires accurate prediction of canonical kinetic constants~\cite{Li2022DLKcat,Yu2023UniKP,quan2024clustering}, \eg, $k_\text{cat}$, $K_\text{m}$, and $K_\text{i}$, directly from molecular inputs, so that large candidate libraries can be screened before any wet-lab investment. Yet current machine learning pipelines~\cite{Boorla2025Catpred,Wang2025Catapro,Shen2024EITLEM,wang2025robust} still struggle to move beyond dataset-specific correlations toward mechanism-aware generalization.

A central difficulty lies in how catalysis actually unfolds~\cite{wu2022reshaping, liu2025data, kraut2003challenges, chen2025glycolysis, buller2023nature,markin2021revealing}, as ``\textit{Mutant Enzyme Reaction Mechanism}" in Figure~\ref{fig:fig1}. Early studies regressed kinetic constants from sequence alone~\cite{barber2022stable,mazurenko2019machine}. Subsequent multimodal systems encoded both the enzyme and the small-molecule substrate, typically via Protein Language Model (PLM)~\cite{Rives2021ESM1b, Meier2021ESM1v, Lin2023ESM2, Hayes2025ESM3} or CNN features for the sequence and SMILES/graph fingerprints for the substrate, and trained joint regressors~\cite{Li2022DLKcat, Yu2023UniKP, Wang2025Catapro, Boorla2025Catpred}. Examples include DLKcat~\cite{Li2022DLKcat} for $k_\text{cat}$ from paired representations, UniKP~\cite{Yu2023UniKP} for multi-endpoint prediction with environmental covariates, and CataPro~\cite{Wang2025Catapro}/CatPred~\cite{Boorla2025Catpred} that add stronger PLM features and uncertainty reporting. Despite progress, most pipelines~\cite{Shen2024EITLEM,Wang2025Catapro,zhao2025temporal} fuse the two branches with shallow operations (\eg, concatenation and a single cross-attention layer) and then regress a score, implicitly treating catalysis as static compatibility. In reality, enzymes \emph{first} recognize and position the substrate, \emph{then} adapt pocket geometry~\cite{leonard1982dimensional,luo2025catalytic,li2025dcbk,du2025improving} to stabilize the transition state; both phases jointly determine $k_\text{cat}$, $K_\text{m}$, and $K_\text{i}$.

Transformer-based PLMs offer a principled prior for mechanism-aware modeling. Large-scale self-supervision yields embeddings that reflect evolutionary constraints, fold regularities, long-range contacts, and conserved catalytic motifs~\cite{Meier2021ESM1v,Lin2023ESM2,Hayes2025ESM3,elnaggar2021prottrans,Elnaggar2023Ankh,Alsamkary2025Ankh3,Truong2023Poet}.
ESM families~\cite{Rives2021ESM1b,Meier2021ESM1v,Lin2023ESM2,Hayes2025ESM3} show that masked-token training captures structure-aware signals transferable to function tasks; ProtTrans/ProtT5~\cite{elnaggar2021prottrans,Truong2023Poet} extend coverage with encoder-decoder modeling; Ankh/Ankh3~\cite{Elnaggar2023Ankh, Alsamkary2025Ankh3} introduce protein-specific multi-task objectives that strengthen residue-level priors. These improved kinetic prediction when used as fixed inputs or lightly tuned backbones, and recent evidence supports careful fine-tuning~\cite{schmirler2024fine}. Two gaps remain: PLMs are used passively rather than being explicitly conditioned on the concrete substrate, and naively injecting 3D evidence can dominate training and erode the biochemical semantics learned during pretraining~\cite{xue2022multimodal,zhou2025protclip, Shen2024EITLEM, Wang2025Catapro, Boorla2025Catpred,wang2026task}. It motivates a formulation that conditions the PLM on substrate semantics and pocket geometry in a staged, mechanism-aligned manner while preserving its sequence-derived prior.

In light of this, we address these gaps by framing enzyme kinetics as \emph{staged conditioning} aligned with enzymology: substrate recognition, then conformational adaptation (as shown in Figure~\ref{fig:fig1}). The proposed \textbf{E}nzyme-\textbf{R}eaction \textbf{B}ridging \textbf{A}dapter (ERBA) adapts to the PLM task in two phases. Stage~1, \emph{Molecular Recognition Cross-Attention} (MRCA), conditions the protein representation on the specific substrate to model selectivity and binding specificity. Stage~2, a \emph{Geometry-aware Mixture-of-Experts} (G-MoE), fuses the MRCA output with an active-site geometry encoder and routes each sample to pocket-specialized experts, capturing induced fit and conformational adaptation where catalysis occurs. To maintain adaptation stability, we introduce \emph{Enzyme-Substrate Distribution Alignment} (ESDA), which defines stability as distributional alignment in a reproducing kernel Hilbert space~\cite{zhang2019optimal,zhang2018aligning} via an RBF-kernel maximum mean discrepancy. ESDA aligns the distributions of sequence-only, sequence-substrate, and sequence-substrate-structure representations to the PLM manifold, preserving biochemical semantics while allowing new substrate and geometry information to be expressed.
Meanwhile, a heteroscedastic Gaussian~\cite{kersting2007most} prediction head trained in $\log_{10}$ space models the positivity of kinetic constants and their multiplicative noise.

In addition, we evaluate across three kinetic endpoints and multiple PLM backbones under consistent protocols.
Extensive results demonstrate that shallow fusion underuses PLM priors and fails to capitalize on geometry, whereas ERBA’s staged conditioning yields consistent gains. Improvements hold across backbones of different scales, and cross-dataset tests indicate stronger out-of-distribution generalization, highlighting the benefit of conditioning the PLM first on substrate semantics and then on pocket geometry in a manner faithful to the catalytic mechanism.

In summary, our principal contributions are as follows:
\begin{itemize}
\item We propose a mechanism-aligned formulation of enzyme kinetics as staged conditioning from substrate recognition to pocket-level adaptation.
\item We introduce an ERBA that combines MRCA for sequence-substrate fusion with a  G-MoE that performs pocket-localized adaptation via sparse expert routing.
\item We design an ESDA that aligns the sequence, substrate, and structure at the distribution level, preserving PLM biochemical semantics during multimodal fine-tuning.
\item Evaluated across three kinetic endpoints and multiple PLM backbones, ERBA delivers consistent gains, offering a general interface for adding new modalities and improving out-of-distribution generalization.
\end{itemize}

\section{Related Work}
\subsection{Enzyme Kinetic Parameter Prediction}
Data-driven prediction of enzyme kinetics has progressed from sequence-only models to multimodal pipelines. Early efforts such as DLKcat~\cite{Li2022DLKcat} coupled CNN/GNN encoders for protein sequences and substrate SMILES. Subsequent systems improved generalization by introducing PLM features~\cite{Kroll2023TurNup, Yu2023UniKP, Shen2024EITLEM, Wang2024MPEK, Wang2025Catapro}: TurNup~\cite{Kroll2023TurNup} pairs ESM-1b~\cite{Rives2021ESM1b} with reaction fingerprints in a boosted regressor; UniKP~\cite{Yu2023UniKP}, MPEK~\cite{Wang2024MPEK}, and CataPro~\cite{Wang2025Catapro} use ProtT5~\cite{elnaggar2021prottrans} with SMILES to predict multiple endpoints; EITLEM~\cite{Shen2024EITLEM} adopts ESM-1v~\cite{Meier2021ESM1v} and residue-level attention for mutants. Despite these gains, most pipelines still operate on 1D sequence and 2D chemistry, leaving out the 3D pocket context that strongly influences $k_\text{cat}$, $K_\text{m}$, and $K_\text{i}$~\cite{luo2025catalytic,li2025dcbk,wang2025robust}.

\begin{figure*}[t!]
\centering
\includegraphics[width=1.0\linewidth]{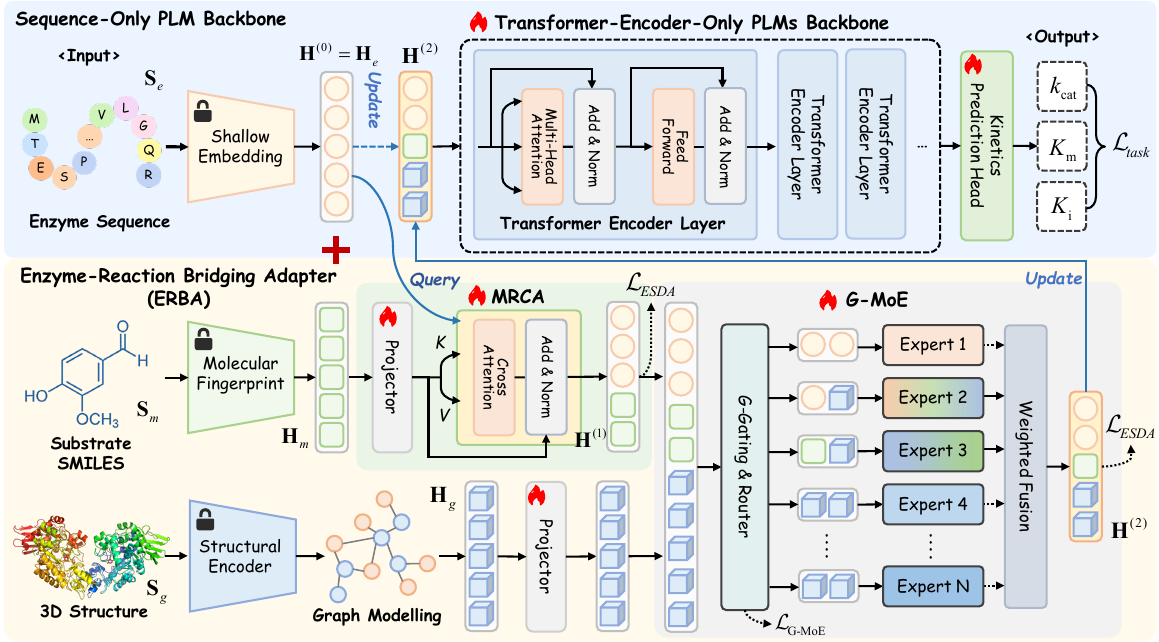}
\vspace{-1.5em}
\caption{\textbf{Architecture of the proposed ERBA.} It augments a sequence-only PLM with multimodal conditioning on substrate chemistry and pocket geometry. MRCA injects substrate fingerprints into enzyme embeddings to capture recognition specificity, and G-MoE integrates local 3D pocket structure to model conformational adaptation. Update and query paths couple both modules to the backbone, while ESDA aligns representations with the prior PLM, enabling accurate prediction of enzyme kinetic parameters.}
\label{fig:overall}
\vspace{-1em}
\end{figure*}
\subsection{Transformer-based Protein Language Models}
Transformer-based PLMs learn contextual protein representations from large sequence corpora~\cite{Meier2021ESM1v, Lin2023ESM2, Hayes2025ESM3, elnaggar2021prottrans, Elnaggar2023Ankh, Alsamkary2025Ankh3, Truong2023Poet,wang2025exploiting}. 
ESM models~\cite{Rives2021ESM1b, Lin2023ESM2} show that masked-token pretraining captures fold regularities, long-range contacts, and mutation sensitivity that transfer to structure and function tasks. 
ProtTrans/ProtT5~\cite{elnaggar2021prottrans,prot5} introduces encoder-decoder training and larger vocabularies, improving robustness for remote homologs. 
Ankh/Ankh3~\cite{Elnaggar2023Ankh,Alsamkary2025Ankh3} adds protein-specific multitask objectives, strengthening residue-level priors. 
Scaling further enriches implicit structural knowledge, as in ESM-3~\cite{Hayes2025ESM3}. 
Such representations are now standard for kinetics because they encode evolutionary constraints, catalytic motifs, and physically plausible contacts. 

\subsection{Multimodal Integration for Enzyme Modeling}
Catalysis reflects coupled effects of sequence, substrate chemistry, and conformational geometry. Recent integrations extend PLMs with structural or chemical modalities~\cite{wang2024multi, hu2025multimodal, ouyang2024mmsite, xiao2025stella, zhou2025protclip}. Examples include multimodal attention for active-site labeling~\cite{wang2024multi}, cross-modal regression with environmental factors~\cite{hu2025multimodal}, and combinations of cryo-EM or structure-derived features with PLM embeddings for molecular property prediction~\cite{yuan2024gpsfun, deng2025chemical, zhou2025protclip, xiao2025stella}. Most works~\cite{Shen2024EITLEM, Wang2025Catapro, Boorla2025Catpred, Wang2024MPEK}, however, fuse modalities shallowly, which undercuts the mechanistic sequence of \emph{recognition} then \emph{adaptation} and risks destabilizing pretrained biochemical priors during fine-tuning. A staged, mechanism-aligned conditioning that injects substrate semantics first and pocket geometry second, while regularizing distributions to remain compatible with PLM space~\cite{bhattacharya2024large,gelman2025biophysics}, is thus warranted.


\section{Methodology}
\subsection{Preliminaries}
\paragraph{Problem Definition.}
The task of predicting enzyme kinetic parameters can be formulated as a conditional regression problem.
Given an enzyme $E$ with amino acid sequence $\mathbf{S}_e=\{s_i\}_{i=1,...,L_e}$, a substrate molecule $\mathbf{S}_m$ (SMILES~\cite{honda2019smiles}) and their structural context $\mathbf{S}_g$, the goal is to predict quantitative kinetic parameters $\hat{\mathbf{y}} = \{k_\text{cat}, K_\text{m}, K_\text{i}\}$.
Formally, we aim to learn a multimodal regression function $f_\theta(\cdot)$ parameterized by~$\theta$:
\begin{equation}
\hat{\mathbf{y}} = f_\theta(\mathbf{S}_e, \mathbf{S}_m, \mathbf{S}_g), \quad 
\hat{\mathbf{y}} \in \mathbb{R}.
\end{equation}
\paragraph{Existing Formulation.}
Most existing methods~\cite{Shen2024EITLEM,Wang2025Catapro,Boorla2025Catpred,Wang2024MPEK} approximate $f_\theta$ by independently encoding the enzyme and substrate, followed by a shallow fusion and regression head to approximate:
\begin{equation}
\hat{\mathbf{y}} = \psi(\mathbf{S}_e \oplus \mathbf{S}_m \oplus \mathbf{S}_g),
\end{equation}
where $\oplus$ denotes concatenation and $\psi(\cdot)$ maps the fused representation to kinetic outputs.
While straightforward, this performs only coarse feature aggregation and lacks fine-grained semantic alignment between molecular modalities.
It implicitly treats catalysis as a static compatibility problem, ignoring the dynamic and hierarchical nature of enzymatic turnover-comprising (i) \textit{substrate recognition} and (ii) \textit{conformational adaptation} within the active site.

\paragraph{Mechanism-Aligned Formulation.}
We instead model enzymatic catalysis as a \textbf{staged conditional process}, in which kinetic behavior emerges from successive molecular interactions. Given that substrate binding determines selectivity and structural adaptation governs efficiency, we reformulate the prediction function as
\begin{equation}\label{equ:target}
\hat{\mathbf{y}} = f_\theta(\mathbf{S}_e \mid \mathbf{S}_m, \mathbf{S}_g)
= f_\theta^{(2)}\!\big(f_\theta^{(1)}(\mathbf{S}_e, \mathbf{S}_m),\, \mathbf{S}_g\big).
\end{equation}

Here, $f_\theta^{(1)}$ captures substrate-conditioned molecular recognition, 
and $f_\theta^{(2)}$ refines the representation through geometry-aware conformational adaptation. 
To realize this staged conditioning, we introduce the \textbf{Enzyme-Reaction Bridging Adapter (ERBA)} $\mathcal{A}(\cdot)$, 
which augments pretrained PLMs through hierarchical multimodal fusion. 
ERBA integrates two key modules: 
(i) the \textbf{Molecular Recognition Cross-Attention (MRCA)} $\mathcal{M}(\cdot)$ that injects substrate semantics into enzyme representations to model recognition specificity, 
and (ii) the \textbf{Geometry-aware Mixture-of-Experts (G-MoE)} $\mathcal{G}(\cdot)$ that incorporates three-dimensional structural cues to capture adaptive catalytic configurations. 
Formally, the PLM-based ERBA reformulates the staged conditioning in Eq.~\ref{equ:target} as:

\begin{equation}
\hat{\mathbf{y}}=\mathcal{A}\big(\mathbf{S}_e \mid \mathbf{S}_m, \mathbf{S}_g\big)=\underbrace{\mathcal{G}^{(2)}\!\big(
\underbrace{\mathcal{M}^{(1)}(\mathbf{S}_e, \mathbf{S}_m)}_{\text{(i) Substrate Recognition}}, \mathbf{S}_g\big)}_{\text{(ii) Conformational Adaptation}},
\end{equation}
where $\mathcal{M}^{(1)}$ performs substrate-conditioned fusion within the PLM latent space to align enzyme and substrate semantics, 
while $\mathcal{G}^{(2)}$ leverages local structural geometry to guide conformation-aware refinement. 
This hierarchical fusion mechanism enables biologically grounded and semantically aligned prediction of enzyme kinetic parameters, 
while preserving the biochemical priors inherent in pretrained PLMs.

\subsection{Molecular Recognition Cross-Attention}
Catalysis begins with molecular recognition: substrate identity constrains selectivity and sets the context for subsequent conformational change.
Let the enzyme sequence be $\mathbf{S}_e\!\in\!\mathbb{R}^{L_e}$ and the substrate string be $\mathbf{S}_m\!\in\!\mathbb{R}^{L_m}$.
Shallow Transformer layers of the pretrained PLM act as an encoder that maps $\mathbf{S}_e$ to residue embeddings $\mathbf{H}_e\in\mathbb{R}^{L_e\times D}$, preserving local biochemical cues while remaining amenable to multimodal conditioning.
And the substrate tokens are embedded by MPNN encoder~\cite{yang2019analyzing} guided by SMILES fingerprint priors~\cite{Boorla2025Catpred}, yielding $\mathbf{H}_m\in\mathbb{R}^{L_m\times D}$.
Recognition is cast as the proposed \textit{Molecular Recognition Cross-Attention} (MRCA) module, which injects substrate semantics into the enzyme environment based on a single-layer cross-attention mechanism. With trainable projections $\mathbf{W}_Q,\mathbf{W}_K,\mathbf{W}_V\in\mathbb{R}^{D\times d_k}$, we compute
\begin{equation}
\begin{aligned}
\mathbf{A}_{em}=&\operatorname{Softmax}\!\Big(\tfrac{(\mathbf{H}_e\mathbf{W}_Q)(\mathbf{H}_m\mathbf{W}_K)^{\!\top}}{\sqrt{d_k}}\Big),\\
\mathbf{Z}_{em}=&\mathbf{A}_{em}(\mathbf{H}_m\mathbf{W}_V),
\end{aligned}
\end{equation}
where $\mathbf{A}_{em}\in\mathbb{R}^{L_e\times L_m}$ aligns enzyme tokens to substrate tokens and aggregates substrate-conditioned evidence.
The resulting update $\mathbf{Z}_{em}\in\mathbb{R}^{L_e\times D}$ is fused back to the enzyme stream with a residual connection and layer normalization, producing the substrate-aware representation $\mathbf{H}^{(1)}\in\mathbb{R}^{L_e\times D}$. This recognition-level summary highlights substrate-relevant residues and is passed to the next stage.

\subsection{Geometry-aware Mixture-of-Experts}
\begin{figure}[t!]
\centering
\includegraphics[width=1.0\linewidth]{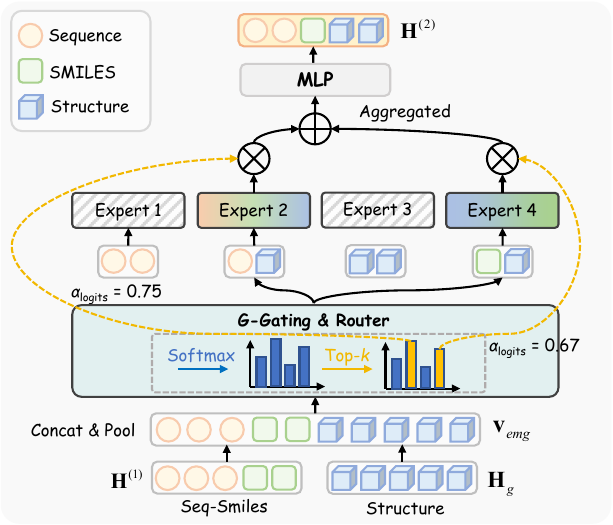}
\vspace{-1.5em}
\caption{G-Gating \& Router pools sequence-substrate and structural cues to produce gating logits, activate the top-\textit{k} geometry-relevant experts, and suppress others. Selected experts transform structure-conditioned features, and an MLP fuses them into a geometry-adaptive representation for kinetic regression.
} 
\label{fig:moe}
\vspace{-1.3 em}
\end{figure}

The second stage conditions recognition features on three-dimensional pocket geometry to model conformational adaptation at the catalytic site. Since pocket topology and residue packing induce heterogeneous geometric regimes that a shared adapter cannot capture~\cite{Boorla2025Catpred}, we introduce a \textit{Geometry-aware Mixture-of-Experts} (G-MoE), whose routing depends jointly on recognition features and pocket structure in Figure~\ref{fig:moe}.
Let $\mathbf{H}^{(1)}\!\in\!\mathbb{R}^{L_e\times D}$ denote the MRCA output and $\mathcal{P}\subset{1,\ldots,L_e}$ the pocket index set. Active-site tokens are encoded by an E-GNN~\cite{greener2025fast} to yield geometry descriptors $\mathbf{H}_g\!\in\!\mathbb{R}^{L_g\times D}$. We construct a joint routing vector $\mathbf{v}_{emg}$ via the G-Gating \& Router module as:


\begin{equation}
\mathbf{v}_{emg}=\Big[\;\operatorname{Pool}\big(\mathbf{H}^{(1)}[\mathcal{P}]\big)\oplus\operatorname{Pool}(\mathbf{H}_g)\;\Big]\in\mathbb{R}^{2D},
\end{equation}
where $\operatorname{Pool}(\cdot)$ is tokenwise mean and $[\mathcal{P}]$ selects the pocket rows. Routing logits $\boldsymbol{\alpha}$ and sparse gates $\widetilde{\boldsymbol{\alpha}}$ are then
\begin{equation}
\begin{aligned}
\boldsymbol{\alpha}&=\operatorname{softmax}(\mathbf{W}_{\text{gate}}\mathbf{v}_{emg}+\mathbf{b}_{\text{gate}})\in\mathbb{R}^{n},\\
\widetilde{\boldsymbol{\alpha}}&=\operatorname{Top-}k(\boldsymbol{\alpha},k)\in\mathbb{R}^{n},
\end{aligned}
\end{equation}
where only the $k$ geometry-relevant experts are activated.
In the model composed of $n$ experts, each expert $E_n$ performs a \emph{pocket-local}, \emph{geometry-modulated low-rank adaptation} by a geometry summary $\Gamma(\mathbf{H}_g)=\operatorname{Pool}(\mathbf{H}_g)\in\mathbb{R}^{D}$:
\begin{equation}
E_n(\mathbf{H}^{(1)}, \mathbf{H}_g) = \mathbf{H}^{(1)} + \mathbf{V}_n \sigma\left(\mathbf{U}_n \mathbf{H}^{(1)}[\mathcal{P}] + \mathbf{B}_n \Gamma(\mathbf{H}_g)\right),
\end{equation}
with $\mathbf{U}_n \in \mathbb{R}^{r \times D}$, $\mathbf{V}_n \in \mathbb{R}^{D \times r}$, and $\mathbf{B}_n\in\mathbb{R}^{r\times D}$ are learnable weight matrices, GELU nonlinearity $\sigma$, and $r\ll D$ is the rank of the adapter.
Geometry thus conditions channel excitation within the pocket while leaving non-pocket residues unchanged, thereby aligning the learned adaptation with the structural determinants of the transition state.

Selected experts are aggregated to obtain the geometry-conditioned state used for kinetic regression as:
\begin{equation}
\mathbf{H}^{(2)}=
\operatorname{MLP}\!\Big(
\sum_{n\in\operatorname{Top}k}\widetilde{\alpha}\,E_n(\mathbf{H}^{(1)},\mathbf{H}_g)
\Big)\in\mathbb{R}^{D}.
\end{equation}

To prevent expert collapse while keeping geometry-driven sparsity, we add a light balancing term $\mathcal{L}_{\text{G-MoE}}$ that nudges the average gate mass toward uniform importance and the usage rate toward the expected Top-$k$ proportion:
\begin{equation}
\mathcal{L}_{\text{G-MoE}}
=\big\|\,\bar{\boldsymbol{\alpha}}-\tfrac{1}{n}\mathbf{1}\,\big\|_2^2
+\big\|\,\bar{\mathbf{n}}-\tfrac{k}{n}\mathbf{1}\,\big\|_2^2,
\end{equation}
where $\bar{\boldsymbol{\alpha}}\in\mathbb{R}^{n}$ is the per-expert average of $\widetilde{\boldsymbol{\alpha}}$ over the current mini-batch, $\bar{\mathbf{u}}\in\mathbb{R}^{n}$ is the per-expert average of the indicator $\mathbf{1}[\widetilde{\boldsymbol{\alpha}}>0]$ over the current mini-batch, $\mathbf{1}\in\mathbb{R}^{n}$ denotes the all-ones vector, and $\|\cdot\|_2$ is the Euclidean norm. This regularizer is lightweight and does not dilute geometry guidance. In combination, geometry-driven routing, pocket-local adaptation, and sparse aggregation yield $\mathbf{H}^{(2)}$ that preserves global PLM priors, incorporates structural evidence at the catalytic site, and improves predictive fidelity for enzyme kinetics.

\begin{table*}[t!]
\setlength\tabcolsep{3pt}
\centering
\resizebox{\textwidth}{!}{
\begin{tabular}{l|c| ccc |cccc |cccc |cccc}
\multirow{2}{*}{Method} & \multirow{2}{*}{Venue} & \multicolumn{3}{c|}{Modality} & 
\multicolumn{4}{c|}{Kinetic Endpoint: $k_\text{cat}$} & \multicolumn{4}{c|}{Kinetic Endpoint: $K_\text{m}$} & \multicolumn{4}{c}{Kinetic Endpoint: $K_\text{i}$} \\
& & $\mathbf{S}_e$ & $\mathbf{S}_m$ & $\mathbf{S}_g$ & R$^2$↑ & PCC↑ & RMSE↓ & MAE↓ & R$^2$↑ & PCC↑ & RMSE↓ & MAE↓ & R$^2$↑ & PCC↑ & RMSE↓ & MAE↓ \\
\thickhline
\multicolumn{17}{l}{\textcolor{gray}{{\textit{\textbf{Exp I: Existing SOTA Models}}}}}\\ 
DLKcat~\cite{Li2022DLKcat}  & Nat.Catal 2022 & \cmark & \cmark &  - & 0.01 & 0.37 & 1.78 & 1.30 & - & - & - & - & - & - & - & - \\
TurNup~\cite{Kroll2023TurNup}  & Nat.C 2023& \cmark & \cmark &  - & 0.39 & 0.65 & 1.60 & 1.19 & - & - & - & - & - & - & - & - \\
UniKP~\cite{Yu2023UniKP}   & Nat.C 2023 & \cmark & \cmark &  - & 0.13 & 0.42 & 1.56 & 1.14 & 0.36 & 0.63 & 1.02 & 0.78 & - & - & - & - \\
MPEK~\cite{Wang2024MPEK}    & Brief.Bio 2024 & \cmark & \cmark &  - & 0.33 & 0.59 & 1.34 & 1.05 & 0.30 & 0.56 & 1.40 & 1.09 & - & - & - & - \\
EITLEM~\cite{Shen2024EITLEM}  & Chem.Catal 2024  & \cmark & \cmark &  - & 0.34 & 0.63 & 1.35 & 0.80 & 0.30 & 0.59 & 1.06 & 0.82 & 0.38 & \secondcolor{0.70} & \secondcolor{1.33} & \secondcolor{0.97} \\
CataPro~\cite{Wang2025Catapro} & Nat.C 2025 & \cmark & \cmark &  - & \secondcolor{0.41} & 0.66 & 1.33 & 0.80 & 0.41 & 0.63 & 1.01 & 0.75 & - & - & - & - \\
CatPred~\cite{Boorla2025Catpred} & Nat.C 2025 & \cmark & \cmark &  \cmark & 0.40 & \secondcolor{0.67} & \secondcolor{1.30} & \secondcolor{0.79} & \secondcolor{0.49} & \secondcolor{0.65} & \secondcolor{0.93} & \secondcolor{0.71} & \secondcolor{0.45} & 0.60 & 1.35 & 1.09 \\
\textbf{Ours$^\ddagger$} & - & \cmark & \cmark &  \cmark & \firstcolor{0.54} & \firstcolor{0.74} & \firstcolor{1.13} & \firstcolor{0.74} & \firstcolor{0.61} & \firstcolor{0.79} & \firstcolor{0.70} & \firstcolor{0.61} & \firstcolor{0.61} & \firstcolor{0.78} & \firstcolor{1.26} & \firstcolor{0.91} \\
\hline
\multicolumn{17}{l}{\textcolor{gray}{{\textit{\textbf{Exp II: Fine-tuning PLM Backbones}}}}}\\ 
Ankh3-1.8B~\cite{Alsamkary2025Ankh3}    & Arxiv 2025 & \cmark &  -& - & 0.41 & 0.65 & 1.28 & 0.92 & 0.40 & 0.64 & 0.97 & 0.74 & 0.41 & 0.64 & 1.54 & 1.22 \\
\rowcolor{gray!20}
\qquad \textit{w}/ ERBA   & - & \cmark & \cmark & \cmark & 0.50 & \secondcolor{0.72} & 1.17 & 0.80 & 0.51 & 0.72 & 0.88 & 0.66 & 0.52 & 0.73 & 1.39 & 1.06 \\
Ankh3-5.7B~\cite{Alsamkary2025Ankh3}   & Arxiv 2025 & \cmark & -& - & 0.43 & 0.67 & 1.24 & 0.88 & 0.42 & 0.65 & 0.98 & 0.74 & 0.46 & 0.68 & 1.47 & 1.15 \\
\rowcolor{gray!20}
\qquad \textit{w}/ ERBA   & - & \cmark & \cmark & \cmark & \secondcolor{0.52} & \firstcolor{0.74} & \firstcolor{1.12} & \secondcolor{0.77} & 0.53 & 0.73 & 0.88 & 0.65 & 0.57 & \secondcolor{0.76} & 1.32 & 1.00 \\
ProtT5-3B~\cite{prot5}  & TPAMI 2021 & \cmark & - & - & 0.39 & 0.65 & 1.30 & 0.91 & 0.39 & 0.63 & 0.98 & 0.74 & 0.43 & 0.66 & 1.50 & 1.14 \\
\rowcolor{gray!20}
\qquad \textit{w}/ ERBA   & - & \cmark & \cmark & \cmark & 0.47 & 0.71 & 1.22 & 0.85 & 0.54 & \secondcolor{0.74} & 0.88 & 0.65 & \secondcolor{0.58} & \secondcolor{0.76} & \secondcolor{1.31} & \secondcolor{0.97} \\
ESM2-8M~\cite{Lin2023ESM2} & Science 2023 & \cmark & - & - & 0.19 & 0.43 & 1.49 & 1.11 & 0.26 & 0.51 & 1.08 & 0.84 & 0.15 & 0.39 & 1.82 & 1.46 \\
\rowcolor{gray!20}
\qquad \textit{w}/ ERBA & - & \cmark & \cmark & \cmark & 0.27 & 0.52 & 1.41 & 1.05 & 0.35 & 0.60 & 1.04 & 0.80 & 0.31 & 0.56 & 1.64 & 1.30 \\
ESM2-35M~\cite{Lin2023ESM2}   & Science 2023& \cmark & - &- & 0.26 & 0.52 & 1.42 & 1.04 & 0.30 & 0.55 & 1.07 & 0.84 & 0.17 & 0.42 & 1.51 & 1.12 \\
\rowcolor{gray!20}
\qquad \textit{w}/ ERBA & - & \cmark & \cmark & \cmark & 0.38 & 0.62 & 1.31 & 0.96 & 0.38 & 0.62 & 1.01 & 0.78 & 0.34 & 0.59 & 1.60 & 1.26 \\
ESM2-150M~\cite{Lin2023ESM2}  & Science 2023 & \cmark & - & - & 0.30 & 0.55 & 1.38 & 1.01 & 0.33 & 0.58 & 1.04 & 0.80 & 0.33 & 0.58 & 1.62 & 1.29 \\
\rowcolor{gray!20}
\qquad \textit{w}/ ERBA & - & \cmark & \cmark & \cmark & 0.38 & 0.62 & 1.31 & 0.95 & 0.54 & 0.73 & 0.88 & 0.65 & 0.53 & 0.73 & 1.36 & 1.02 \\
ESM2-650M~\cite{Lin2023ESM2}  & Science 2023 & \cmark & - & -& 0.35 & 0.59 & 1.33 & 0.98 & 0.39 & 0.63 & 0.97 & 0.75 & 0.36 & 0.60 & 1.58 & 1.27 \\
\rowcolor{gray!20}
\qquad \textit{w}/ ERBA & - & \cmark & \cmark & \cmark & 0.44 & 0.66 & 1.25 & 0.91 & \secondcolor{0.55} & \secondcolor{0.74} & \secondcolor{0.81} & \secondcolor{0.64} & 0.53 & 0.74 & 1.35 & 0.99 \\
ESM2-3B~\cite{Lin2023ESM2}    & Science 2023 & \cmark & - & - & 0.41 & 0.65 & 1.27 & 0.92 & 0.43 & 0.66 & 0.94 & 0.72 & 0.49 & 0.71 & 1.38 & 1.05 \\
\rowcolor{gray!20}
\textbf{\textit{w}/ ERBA (Ours)} & - & \cmark & \cmark & \cmark & \firstcolor{0.54} & \firstcolor{0.74} & \secondcolor{1.13} & \firstcolor{0.74} & \firstcolor{0.61} & \firstcolor{0.79} & \firstcolor{0.70} & \firstcolor{0.61} & \firstcolor{0.61} & \firstcolor{0.78} & \firstcolor{1.26} & \firstcolor{0.91} \\
\end{tabular}
}
\vspace{-1 em}
\caption{\textbf{Performance comparison of existing SOTA methods and fine-tuned PLM backbones on enzyme kinetic prediction tasks.} Results are presented for three kinetic parameters: turnover number ($k_{\text{cat}}$), Michaelis constant ($K_\text{m}$), and inhibition constant ($K_\text{i}$). The performance metrics include R$^2$↑, PCC↑, RMSE↓, and MAE↓. ERBA consistently outperforms the baselines across all metrics. 
Notably, the best value in each column is highlighted in \colorbox{firstcolor}{red}, and the second best in \colorbox{secondcolor}{blue}. $^\ddagger$ indicates that the backbone of the PLMs is fine-tuned.
}
\vspace{-1.5em}
\label{tab:exp}
\end{table*}

\subsection{Enzyme-Substrate Distribution Alignment}
Staged conditioning should absorb substrate and geometry signals while remaining compatible with the biochemical prior encoded by the pretrained PLM. To keep both stages anchored to this prior, let $\mathbf{H}^{(0)}=\mathbf{H}_e$ is the sequence embedding, $\mathbf{H}^{(1)}$ the substrate-conditioned embedding from MRCA, and $\mathbf{H}^{(2)}$ the geometry-conditioned representation from G\mbox{-}MoE. We introduce the \emph{Enzyme-Substrate Distribution Alignment} (ESDA), inspired by kernel-based two-sample tests and distribution alignment techniques in statistical learning and transfer adaptation~\cite{lv2022debiased, guo2023identifying, wang2025pre}. ESDA formalizes stability as distributional alignment in a reproducing kernel Hilbert space~\cite{zhang2019optimal,zhang2018aligning}, ensuring that recognition and adaptation evolve within the semantic domain of the PLM, rather than drifting into a new manifold.
For this alignment, each stage’s global representation is summarized by averaging the embeddings across residues:
\begin{equation}
\mathcal{Z}^{(t)}=\frac{1}{L_e}\sum_{i=1}^{L_e}\mathbf{H}^{(t)}[i],\quad t\in\{0,1,2\}.
\end{equation}

For any indices $a, b\!\in\!t$, using an RBF kernel $\kappa(\mathcal{Z}^{a},\mathcal{Z}^{b})=\exp\!\big(-\|\mathcal{Z}^{a}-\mathcal{Z}^{b}\|_2^2/(2\sigma^2)\big)$ with bandwidth $\sigma$ set by the median heuristic~\cite{jayasumana2015kernel}, we adopt a diagonal-free, normalized mini-batch Maximum Mean Discrepancy (MMD) estimator~\cite{yan2017mind} for any two sets $\mathcal{Z}^{a},\mathcal{Z}^{b}$ of sizes $N_a,N_b$:
\begin{equation}
\begin{aligned}
\widehat{\mathrm{MMD}}^{2}(\mathcal{Z}^{a},\mathcal{Z}^{b})
=&\frac{1}{N_a^{2}}\!\!\sum_{\substack{p,q\\ p\neq q}}\kappa(\mathcal{Z}^{a}_{p},\mathcal{Z}^{a}_{q})
+\frac{1}{N_b^{2}}\!\!\sum_{\substack{p,q\\ p\neq q}}\kappa(\mathcal{Z}^{b}_{p},\mathcal{Z}^{b}_{q})\\
&-\frac{2}{N_aN_b}\sum_{p,q}\kappa(\mathcal{Z}^{a}_{p},\mathcal{Z}^{b}_{q}).
\end{aligned}
\end{equation}

In summary, ESDA anchors both stages to the PLM manifold and encourages a smooth progression from recognition to adaptation:
\begin{equation}
\mathcal{L}_{\mathrm{ESDA}}
=\widehat{\mathrm{MMD}}^{2}\!\big(\mathcal{Z}^{(1)},\mathcal{Z}^{(0)}\big)
+\widehat{\mathrm{MMD}}^{2}\!\big(\mathcal{Z}^{(2)},\mathcal{Z}^{(0)}\big).
\end{equation}

By aligning distributions rather than individual features, ESDA suppresses feature flooding from high-capacity geometric channels and mitigates forgetting of PLM biochemical semantics, while retaining information introduced by substrate identity and pocket geometry.

\subsection{Enzyme Reaction Optimization}
Kinetic constants are positive and affected by multiplicative noise, so we follow~\cite{Boorla2025Catpred} to regress in $\log_{10}$ space using a heteroscedastic Gaussian negative log-likelihood (NLL)~\cite{germain2016pac}.
For a target $y>0$, let $z=\log_{10}y$. The prediction head outputs a mean $\mu$ and a log-variance $s=\log\sigma^{2}$ for $z$, yielding the per-sample NLL:
\begin{equation}
\label{eq:nll_log10_compact}
\mathcal{L}_{\text{task}}
=\frac{1}{2}\,e^{-s}\,(z-\mu)^{2}+\frac{1}{2}\,s.
\end{equation}

Therefore, the full objective $\mathcal{L}$ aggregates the likelihood through hyperparameter ($\lambda_1, \lambda_2$) modulation of task regularization $\mathcal{L}_{\text{task}}$, geometry-aware specialization $\mathcal{L}_{\text{G-MoE}}$, and distributional alignment $\mathcal{L}_{\text{ESDA}}$:
\begin{equation}
\label{eq:total}
\mathcal{L}
=\mathcal{L}_{\text{task}}
+\lambda_1\,\mathcal{L}_{\text{G-MoE}}
+\lambda_2\,\mathcal{L}_{\text{ESDA}}.
\end{equation}

\section{Experiments}

\begin{table*}[t!]
\setlength\tabcolsep{3pt}
\centering
\resizebox{\textwidth}{!}{
\begin{tabular}{l|c|G|GGGG |H|HHHH |I|IIII}
\multirow{2}{*}{Index} & \multirow{2}{*}{Class Name} &
\multicolumn{5}{c|}{Kinetic Endpoint: $k_\text{cat}$} & \multicolumn{5}{c|}{Kinetic Endpoint: $K_\text{m}$} & \multicolumn{5}{c}{Kinetic Endpoint: $K_\text{i}$} \\
& & 
\cellcolor{white}Count & \cellcolor{white}R$^2$\(\uparrow\) & \cellcolor{white}PCC\(\uparrow\) & \cellcolor{white}RMSE\(\downarrow\) & \cellcolor{white}MAE\(\downarrow\) &
\cellcolor{white}Count & \cellcolor{white}R$^2$\(\uparrow\) & \cellcolor{white}PCC\(\uparrow\) & \cellcolor{white}RMSE\(\downarrow\) & \cellcolor{white}MAE\(\downarrow\) &
\cellcolor{white}Count & \cellcolor{white}R$^2$\(\uparrow\) & \cellcolor{white}PCC\(\uparrow\) & \cellcolor{white}RMSE\(\downarrow\) & \cellcolor{white}MAE\(\downarrow\) \\
\thickhline
EC-1 & \textit{Oxidoreductases} & 679 & 0.37 & 0.62 & 1.32 & 0.95 & 1,382 & 0.55 & 0.74 & 0.90 & 0.69 & 282 & 0.58 & 0.77 & \firstcolor{1.19} & \secondcolor{0.92} \\
EC-2 & \textit{Transferases} & 334 & 0.46 & 0.69 & 1.22 & \secondcolor{0.87} & 1,000 & 0.53 & 0.73 & \firstcolor{0.81} & \firstcolor{0.61} & 270 & 0.38 & 0.65 & 1.37 & 0.99 \\
EC-3 & \textit{Hydrolases} & 682 & 0.43 & 0.68 & \secondcolor{1.18} & 0.88 & 1,145 & \secondcolor{0.51} & 0.72 & 0.90 & 0.65 & 408 & 0.48 & 0.71 & 1.50 & 1.10 \\
EC-4 & \textit{Lyases} & 131 & \secondcolor{0.50} & 0.71 & 1.27 & 0.92 & 293   & 0.56 & \secondcolor{0.75} & \firstcolor{0.81} & \firstcolor{0.61} & 157 & \secondcolor{0.65} & \secondcolor{0.81} & \firstcolor{1.19} & \firstcolor{0.84} \\
EC-5 & \textit{Isomerases} &  72 & 0.46 & \secondcolor{0.73} & 1.39 & 0.94 & 123   & 0.61 & \firstcolor{0.78} & \secondcolor{0.84} & \secondcolor{0.64} &  42 & \firstcolor{0.67} & \firstcolor{0.83} & \secondcolor{1.20} & 0.95 \\
EC-6 & \textit{Ligases} &  48 & \firstcolor{0.53} & \firstcolor{0.77} & \firstcolor{0.96} & \firstcolor{0.71} & 135 & \firstcolor{0.34} & 0.59 & 1.02 & 0.76 &  21 & 0.34 & 0.62 & 1.39 & 1.03 \\
\end{tabular}
}
\caption{\textbf{Enzyme category analysis across kinetic prediction tasks.} The table provides a breakdown of enzyme categories in the datasets for $k_{\text{cat}}$, $K_\text{m}$, and $K_\text{i}$ prediction tasks. It includes the distribution of each enzyme class, the number of samples per category, and performance metrics for each class.
Notably, the best value in each column is highlighted in \colorbox{firstcolor}{red}, and the second best in \colorbox{secondcolor}{blue}.
}
\label{tab:class}
\vspace{-1.2 em}
\end{table*}

\begin{table}[t!]
\setlength\tabcolsep{3pt}
\centering
\resizebox{\columnwidth}{!}{
\begin{tabular}{l|cccc |cccc}
\multirow{2}{*}{Method} & \multicolumn{4}{c|}{\textbf{\textit{OOD Test @EITLEM-$k_\text{cat}$}}} & \multicolumn{4}{c}{\textbf{\textit{OOD Test @EITLEM-$K_\text{m}$}}} \\
 & R$^2$↑ & PCC↑ & RMSE↓ & MAE↓ & R$^2$↑ & PCC↑ & RMSE↓ & MAE↓\\
\thickhline
DLKcat~\cite{Li2022DLKcat} & 0.01 & 0.15 & 3.87 & 3.05 & - & - & - & - \\
TurNup~\cite{Kroll2023TurNup} & 0.21 & 0.38 & 2.46 & 2.11 & - & - & - & - \\
UniKP~\cite{Yu2023UniKP}  & 0.08 & 0.20 & 1.76 & 1.24 & 0.14 & 0.32 & 1.54 & 1.08  \\
MPEK~\cite{Wang2024MPEK} & 0.17 & 0.41 & 1.65 & 1.18 & 0.12 & 0.35 & 1.71 & 1.32  \\
EITLEM~\cite{Shen2024EITLEM} & \secondcolor{0.27} & \secondcolor{0.50} & \secondcolor{1.43} & \secondcolor{0.95} & 0.23 & 0.43 & 1.31 & 1.03  \\
CataPro~\cite{Wang2025Catapro}& 0.25 & 0.47 & 1.59 & 1.08 & 0.27 & 0.41 & 1.23 & 0.96 \\
CatPred~\cite{Boorla2025Catpred} & 0.25 & 0.46 & 1.50 & 1.03 & \secondcolor{0.30} & \secondcolor{0.45} & \secondcolor{1.11} & \secondcolor{0.87}  \\
\textbf{Ours$^\ddagger$}& \firstcolor{0.50} & \firstcolor{0.70} & \firstcolor{1.21} & \firstcolor{0.81} & \firstcolor{0.55} & \firstcolor{0.69} & \firstcolor{0.79}& \firstcolor{0.69} \\
\end{tabular}
}
\caption{\textbf{Comparison with existing SOTA models on OOD tests.} $k_\text{cat}$ and $K_\text{m}$ using the EITLEM~\cite{Shen2024EITLEM} test sets.
}
\label{tab:ood}
\vspace{-1.2 em}
\end{table}
\subsection{Experimental Setup}
\paragraph{Datasets.}
Following prior works~\cite{Boorla2025Catpred, Shen2024EITLEM}, we evaluate on three standard kinetic parameter prediction tasks: $k_{\text{cat}}$, $K_{\text{m}}$, and $K_{\text{i}}$, using data from BRENDA~\cite{schomburg2004brenda} and SABIO-RK~\cite{wittig2012sabio}. The datasets contain $23{,}197$ records for $k_{\text{cat}}$, $41{,}174$ for $K_{\text{m}}$, and $11{,}929$ for $K_{\text{i}}$. Each entry includes a UniProt-validated enzyme sequence, substrate SMILES string, experimentally measured kinetic value, and associated 3D structure annotation. Compared to datasets used in DLKcat~\cite{Li2022DLKcat} and UniKP~\cite{Yu2023UniKP}, we extend sequence and EC coverage, standardize molecular representations, and reduce integration biases, enabling controlled multimodal evaluation. Notably, in Section~\ref{ood}, we use the $k_\text{cat}$ and $K_\text{m}$ test sets from~\cite{Shen2024EITLEM} for out-of-distribution generalization testing, where 3D structures are generated by OpenFold~\cite{ahdritz2024openfold} or ESMFold~\cite{Lin2023ESM2}. Details are given in the Appendix.  

\paragraph{Evaluation Metrics.}
We evaluate using mean absolute error (MAE)~\cite{Li2022DLKcat, wang2024eulermormer}, root mean square error (RMSE)~\cite{Wang2025Catapro,wang2024frequency}, Pearson correlation coefficient (PCC)~\cite{Boorla2025Catpred}, and coefficient of determination (R$^2$)~\cite{Li2022DLKcat,Boorla2025Catpred,Shen2024EITLEM}. All metrics are computed on $\log_{10}$ transformed targets for $k_{\text{cat}}$, $K_{\text{m}}$, and $K_{\text{i}}$ to stabilize scale and variance. Lower MAE and RMSE and higher PCC and R$^2$ indicate better performance. Unless noted otherwise, results are averaged over repeated runs with fixed data splits and identical preprocessing.

\paragraph{Implementation Details.}
The enzyme-encoding process employs a PLM with a latent dimension of $D=1024$ and a fixed tokenization scheme. Sequences are truncated or padded to a maximum length $L_e=1024$. The substrate is encoded from a canonical SMILES and mapped to dimension $D$ via a linear layer; pocket geometry is computed from active-site residues, and its channels are projected to $D$ via another linear layer. G\text{-}MoE employs $n=4$ experts with top $k=2$ routing. We evaluate ProtT5~\cite{prot5}, ESM2~\cite{Lin2023ESM2}, and Ankh3~\cite{Alsamkary2025Ankh3} backbones to verify robustness across PLMs. For parameter-efficient adaptation, LoRA~\cite{hu2022lora} is applied to the top PLM layers with a rank of $8$, a scaling factor of $16$, and a dropout rate of $0.1$. Training uses AdamW~\cite{loshchilov2017decoupled} with learning rate $1\times10^{-4}$. In full objective $\mathcal{L}$, the hyperparameters are set to $\lambda_{1}=0.01$ and $\lambda_{2}=0.1$.


\subsection{Performance Analysis}
We benchmark ours (ESM2-3B \textit{w}/ ERBA) against recent methods on $k_{\text{cat}}$, $K_{\text{m}}$, and $K_{\text{i}}$, reimplementing all baselines with the same ESM2-3B enzyme encoder for fairness.

\paragraph{Performance Comparison.}
As shown in Exp I of Table~\ref{tab:exp}, our method outperforms existing SOTA models across all three kinetic parameters. For $k_{\text{cat}}$, we achieve an $R^2$ of 0.54, a notable improvement over CatPro's 0.41, along with superior PCC (0.74 \textit{vs.} 0.67) and lower RMSE (0.79 \textit{vs.} 1.30). For $K_\text{m}$, our model leads with an R$^2$ of 0.61, compared to CatPro's 0.49, and a lower RMSE of 0.70. Similarly, for $K_\text{i}$, we achieve a PCC of 0.78 and reduce RMSE to 1.26, outperforming CatPro's 0.60 and 1.35, respectively.
These results underscore the effectiveness of our approach, which combines sequence and substrate information via ERBA to enable accurate modeling of enzyme-substrate interactions.
\begin{figure*}[t!]
\centering
\includegraphics[width=0.95\linewidth]{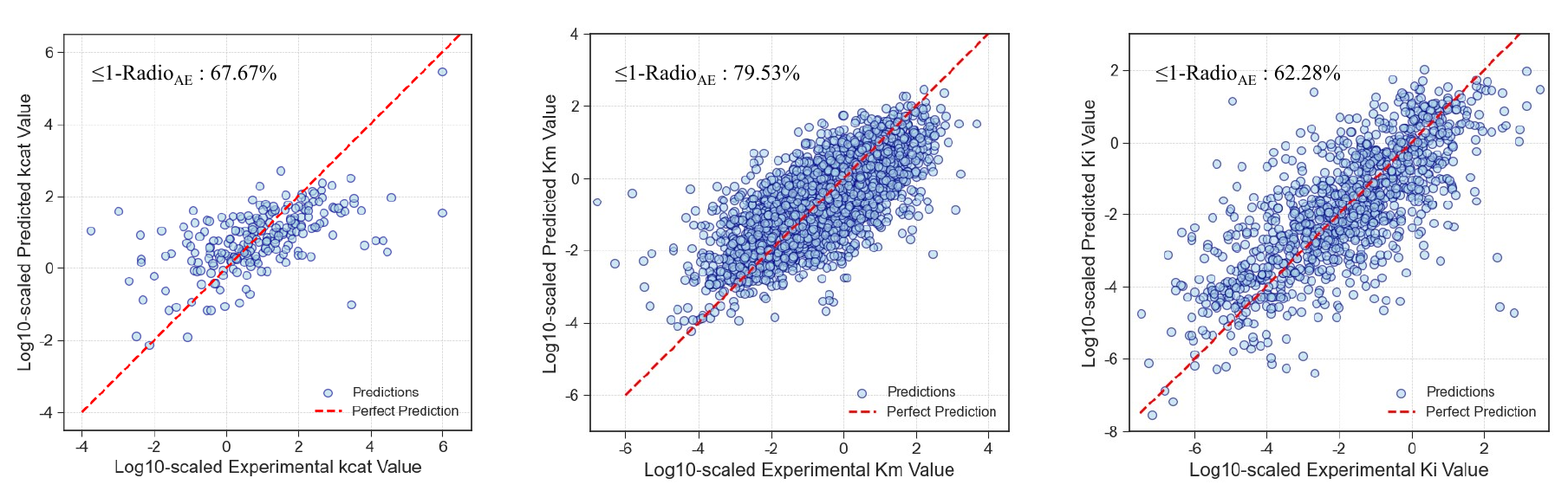}
\vspace{-1.5 em}
\caption{
\textbf{Log-scaled experimental versus predicted values for the kinetic parameters} $k_{\text{cat}}$\textbf{,} $K_\text{m}$\textbf{, and} $K_\text{i}$.
Each plot shows the absolute error less than or equal to 1 as a percentage, denoted as 1-$\text{Radio}_{\text{AE}}$. The dashed red line represents perfect predictions. 
} 
\label{fig:ae}
\vspace{-1.3 em}
\end{figure*}
\begin{figure}[t!]
\centering
\includegraphics[width=1.0\linewidth]{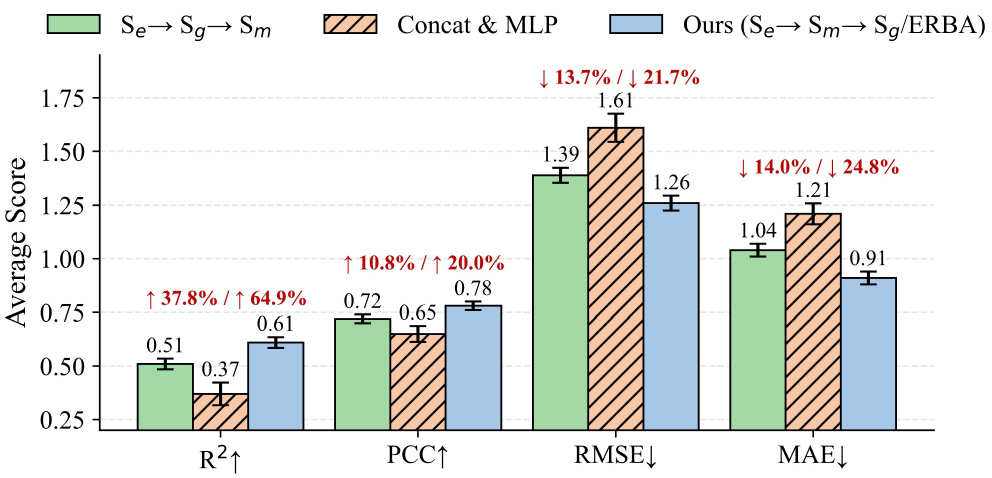}
\vspace{-1.5 em}
\caption{\textbf{Ablation studies on fusion order and manner.} Comparison of different fusion strategies: $\mathbf{S}_e$→$\mathbf{S}_g$→$\mathbf{S}_m$, Concat \& MLP, and the proposed $\mathbf{S}_e$→$\mathbf{S}_m$→$\mathbf{S}_g$/ERBA. Percentage improvements across metrics are highlighted in {\color{red}red}.
}
\label{fig:ab_order}
\vspace{-1.2em}
\end{figure}

\paragraph{Backbone Scaling and the Effect of ERBA.}
We next examine whether ERBA’s gains persist across PLM scales and families in Exp II of Table~\ref{tab:exp}. 
Across ESM2 (8\,M $\rightarrow$ 3\,B), ProtT5-3B, and Ankh3 (1.8\,B /5.7\,B), attaching ERBA to the same backbone consistently improves all three endpoints. On ESM2-3B, $k_{\text{cat}}$ rises from R$^2{=}0.41$ to $0.54$ with higher PCC (0.74) and lower RMSE (0.79); $K_\text{m}$ climbs from R$^2{=}0.43$ to $0.61$ with RMSE dropping to $0.70$; $K_\text{i}$ sees PCC improve from 0.71 to 0.78 while RMSE falls from 1.38 to 1.26. Similar trends hold at smaller scales (8\,M /35\,M /150\,M /650\,M) and on non-ESM backbones (ProtT5-3B, Ankh3), indicating that ERBA provides orthogonal benefits to model size and pretraining recipe. Scaling the backbone alone yields moderate gains, but adding ERBA produces larger, more uniform improvements, especially on $K_\text{m}$ and $K_\text{i}$ where structural context matters most.

\paragraph{Performance Representation of Enzyme Distribution.}
Table~\ref{tab:class} presents the performance across six enzyme classes for the kinetic parameters $k_{\text{cat}}$, $K_\text{m}$, and $K_\text{i}$. Our model performs best for EC-4 (Lyases), achieving high R$^2$ and PCC values, particularly for $K_\text{i}$ and $k_{\text{cat}}$. EC-6 (Ligases) shows lower performance, particularly in $K_\text{m}$ and $K_\text{i}$. These results demonstrate the model's overall robustness while highlighting challenges in predicting certain kinetic parameters for specific enzyme classes.

\paragraph{Error Distribution Analysis.}
Figure~\ref{fig:ae} shows the log-scaled experimental \textit{vs.} predicted values for $k_\text{cat}$, $K_\text{m}$, and $K_\text{i}$, with 1-Radio$_\text{AE}$ indicating the percentage of predictions with absolute errors $\leq$1. The dashed red line represents perfect predictions.
Overall, the model demonstrates good performance, with $1$-Radio$\text{AE}$ values of 67.67\% for $k\text{cat}$, 62.28\% for $K_\text{i}$, and 79.53\% for $K_\text{m}$. $K_\text{m}$ predictions are more concentrated around the perfect prediction line, while $k_\text{cat}$ shows more variation.
These results highlight the model's robustness and suggest that further improvement is needed, particularly for $k_\text{cat}$ and $K_\text{i}$.

\paragraph{Out-of-Distribution Generalization.}\label{ood}
To evaluate OOD generalization, we compare our method with SOTA models on $k_\text{cat}$ and $K_\text{m}$ using the EITLEM test sets~\cite{Shen2024EITLEM}. Our model consistently outperforms all other methods. For $k_\text{cat}$, we achieve the highest R$^2$ (0.50) and PCC (0.70), outperforming EITLEM (R$^2$ = 0.27, PCC = 0.50) and others. For $K_\text{m}$, we achieve R$^2$ of 0.55 and PCC of 0.69, surpassing SOTA models by a clear margin. These results show that leveraging PLMs as dynamic backbones, rather than fixed priors, enables superior generalization. Our ERBA adapts the PLM’s pre-trained knowledge to specific substrate and structural contexts, providing enhanced flexibility. This approach improves the model’s ability to generalize across unseen enzyme-substrate pairs, giving it a clear advantage over models that use PLMs passively.

\subsection{Ablation Study}
\paragraph{Do the Modules Add Gains?}
We assess the contribution of each module by progressively adding MRCA, G-MoE, and ESDA to the PLM baseline. In Table~\ref{tab:ab}, MRCA improves R$^2$ from 0.49 to 0.51 and PCC from 0.71 to 0.73, enhancing substrate recognition. Adding G-MoE further refines the model, increasing R$^2$ to 0.54 and PCC to 0.74. Finally, the complete ESDA framework achieves the best performance with R$^2$ of 0.61 and PCC of 0.78, demonstrating the value of distribution alignment for preserving PLM semantics while incorporating substrate and geometry.

\paragraph{Does Fusion Order Matter?}
The results show that the fusion order significantly impacts model performance. Specifically, the $\mathbf{S}_e\!\to\!\mathbf{S}_m\!\to\!\mathbf{S}_g$/ERBA fusion consistently outperforms the $\mathbf{S}_e\!\to\!\mathbf{S}_g\!\to\!\mathbf{S}_m$ configuration across all metrics. This demonstrates that first aligning the enzyme sequence with substrate information before incorporating structural context leads to better model generalization. The observed improvement in R$^2$ (37.8\%) and a reduction in RMSE (20\%) highlight the importance of this fusion order for accurate enzyme kinetic predictions.

\paragraph{Is ESBA Better than Concatenation?}
The comparison between our ESBA and simple concatenation with MLP~\cite{zhang2024llama, zhang2025multimodal} highlights the advantage of incorporating distributional alignment. ESBA improves model performance, yielding a 20\% increase in PCC and a 14\% reduction in MAE compared to concatenation. It demonstrates ESBA by aligning the representations at the distribution level, preserving the pretrained biochemical priors while effectively integrating the substrate and structural information.
\begin{table}[t!]
\centering
\resizebox{\columnwidth}{!}{
\begin{tabular}{c|ccc |cccc }
\multirow{2}{*}{Method} & \multicolumn{3}{c|}{Modality} &  \multicolumn{4}{c}{Kinetic Endpoint: $K_\text{i}$} \\
& $\mathbf{S}_e$ & $\mathbf{S}_m$ & $\mathbf{S}_g$ & R$^2$↑ & PCC↑ & RMSE↓ & MAE↓\\
\thickhline
PLM Baseline &\cmark & - & -& 0.49 &0.71 &1.38 &1.05 \\
\textit{w}/ MRCA & \cmark & \cmark & - &0.51 &0.73 &1.32 &0.98 \\
\textit{w}/ G-MoE & \cmark & \cmark & \cmark &0.54 &0.74 &1.27 &0.94 \\
\rowcolor{gray!20}
\textbf{\textit{w}/ ESDA (Ours)} & \cmark & \cmark & \cmark &\textbf{0.61} &\textbf{0.78} &\textbf{1.26} &\textbf{0.91} \\
\end{tabular}
}
\caption{\textbf{Ablation study on the effect of different modules.} Performance comparison for $K_i$ prediction with various configurations: PLM baseline, MRCA, G-MoE, and our ESDA framework.}
\label{tab:ab}
\vspace{-1.5em}
\end{table}

\section{Conclusion}
We frame enzyme kinetics as mechanism-aligned staged conditioning, not static compatibility, and propose ERBA, which fine-tunes PLMs with MRCA for substrate-conditioned representations and G-MoE for pocket-aware fusion with geometry. ESDA enforces distributional alignment among sequence, sequence-substrate, and sequence-substrate-structure so that new modalities refine rather than overwrite its prior, yielding consistent gains and stronger out-of-distribution generalization, and offering a compact recipe for future multimodal PLMs that tightly couple sequence, molecular graphs, and geometry.

\section{Acknowledgments}
This research was supported by the Natural Science Foundation of Anhui Province (2408085MC051, 2408085MF159), the Key Science \& Technology Project of Anhui Province (202304a05020068, 202423l10050064), the National Natural Science Foundation of China (U24A20331), and the Fundamental Research Funds for the Central Universities of China (PA2025IISL0109, JZ2023HGQA0472). Yanyan Wei is the corresponding author.



{
    \small
    \bibliographystyle{ieeenat_fullname}
    \bibliography{main}
}

\section*{\centering Overview}
The supplementary materials provide additional details and elaborations on the datasets used, quantitative analysis experiments, and ablation studies presented in the main manuscript.
These topics are organized as follows:

\tableofcontents

\begin{table*}[t!]
\setlength\tabcolsep{3pt}
\centering
\resizebox{\textwidth}{!}{
\begin{tabular}{l|ccc|c|m{12cm}}
\toprule[1.2pt]
\multirow{2}{*}{\textbf{Class}} & \multicolumn{3}{c|}{\textbf{Kinetic Dataset}} & \multirow{2}{*}{\textbf{Total}} & \multirow{2}{*}{\textbf{Details}} \\
& \textbf{$\mathit{k}_{\text{cat}}$} & \textbf{$K_\text{m}$} & \textbf{$K_\text{i}$} & & \\
\hline
\multicolumn{6}{l}{\textcolor{gray}{{\textit{\textbf{Dataset I (Main Test): $k_\text{cat}, K_\text{m}$ and $K_\text{i}$ from~\cite{Boorla2025Catpred}}}}}} \\ 
Entries & 23,151 & 41,174 & 11,929 & 76,254 &  \\
\hline
\cellcolor{gray!10}EC 1 Oxidoreductases & 7,756 & 13,931 & 2,744 & 24,431 & Oxidoreductases catalyze diverse redox reactions by transferring electrons between molecules and substrates, and represent the largest EC class with broad kinetic and functional diversity.\\
\hline
\cellcolor{gray!5}EC 2 Transferases & 4,155 & 10,182 & 2,706 & 17,043 & Transferases catalyze the transfer of functional groups between diverse substrates, enabling the formation of new chemical bonds and the rapid and efficient interconversion of molecules.\\
\hline
\cellcolor{gray!10}EC 3 Hydrolases & 8,065 & 11,025 & 4,083 & 23,173 & Hydrolases catalyze the hydrolysis of diverse chemical bonds in biological molecules. They efficiently cleave covalent linkages in various substrates with variable enzyme kinetic parameters. \\
\hline
\cellcolor{gray!5}EC 4 Lyases & 1,620 & 2,962 & 1,525 & 6107 & Lyases catalyze the cleavage of chemical bonds without hydrolysis or oxidation. They form double bonds or ring structures in substrates. \\
\hline
\cellcolor{gray!10}EC 5 Isomerases & 928 & 1,311 & 434 & 2,673 & Isomerases catalyze the conversion of molecules into their isomeric forms by reorganizing atomic connectivity within the same substrate. \\
\hline
\cellcolor{gray!5}EC 6 Ligases & 590 & 1,363 & 254 & 2,207 & Ligases catalyze the formation of covalent bonds between two different molecules using chemical energy from nucleotide triphosphates. \\
\hline
\cellcolor{gray!10}EC 7 Translocases & 16 & 98 & 23 & 137 & Translocases catalyze the active transport of ions or molecules across biological membranes, efficiently coupled with chemical reaction energy. \\
\hline
\cellcolor{gray!5}Unclassified/Invalid & 21 & 302 & 160 & 483 & Entries that are unclassified or contain format errors, excluded from standard EC categorization. \\
\hline
Unique Sequences & 7,183 & 12,355 & 2,829 & 22,367 &  \\ \hline
Unique EC Classes & 2,657 & 3,550 & 1,306 & 7,513 & \\ \hline
Unique Organisms & 1,685 & 2,419 & 652 & 4,756 & \\ \hline
\hline
\multicolumn{6}{l}{\textcolor{gray}{{\textit{\textbf{Dataset II (OOD Test): $k_\text{cat}$ and $K_\text{m}$ from~\cite{Shen2024EITLEM}}}}}}\\ 
Entries & 35,147 & 29,321 & - & 64,468 & - \\
\hline
\cellcolor{gray!10} EC 1 Oxidoreductases & 12,197 & 9,770 & - & 21,967 & Oxidoreductases catalyze diverse redox reactions by transferring electrons between molecules and substrates, and represent the largest EC class with broad kinetic and functional diversity. \\
\hline
\cellcolor{gray!5} EC 2 Transferases & 8,148 & 7,820 & - & 15,968 & Transferases catalyze the transfer of functional groups between diverse substrates, enabling the formation of new chemical bonds and the rapid and efficient interconversion of molecules. \\
\hline
\cellcolor{gray!10} EC 3 Hydrolases & 8,372 & 6,537& -  & 14,909 & Hydrolases catalyze the hydrolysis of diverse chemical bonds in biological molecules. They efficiently cleave covalent linkages in various substrates with variable enzyme kinetic parameters. \\
\hline
\cellcolor{gray!5} EC 4 Lyases & 3,264 & 2,510 & - & 5,774 & Lyases catalyze the cleavage of chemical bonds without hydrolysis or oxidation. They form double bonds or ring structures in substrates. \\
\hline
\cellcolor{gray!10} EC 5 Isomerases & 1,854 & 1,367& -  & 3,221 & Isomerases catalyze the conversion of molecules into their isomeric forms by reorganizing atomic connectivity within the same substrate. \\
\hline
\cellcolor{gray!5} EC 6 Ligases & 1,252 & 1,187 & - & 2,439 & Ligases catalyze the formation of covalent bonds between two different molecules using chemical energy from nucleotide triphosphates. \\
\hline
\cellcolor{gray!10} EC 7 Translocases & 60 & 130 & - & 190 & Translocases catalyze the active transport of ions or molecules across biological membranes, efficiently coupled with chemical reaction energy. \\
\hline
\end{tabular}
}
\caption{Detailed breakdown of the enzyme kinetic datasets from~\cite{Boorla2025Catpred,Shen2024EITLEM} used for training and out-of-distribution (OOD) testing. We provide the number of entries, unique sequences, EC classes, organisms, and additional information for each enzyme class in the main and OOD test datasets.}
\label{tab:sup_dataset}
\vspace{-0.5em}
\end{table*}

\section{Enzyme Kinetic Parameter Dataset}

\subsection{Details Dataset from~\cite{Boorla2025Catpred}}

In the main experiments of our manuscript, we employ three specially curated kinetic endpoint datasets, as shown in Table~\ref{tab:sup_dataset}, to construct and evaluate learning-based models for predicting enzyme kinetic parameters.
The raw data underlying the three endpoint datasets were curated from two authoritative biochemical databases: BRENDA~\cite{schomburg2004brenda} release 2022\_2
and SABIO-RK~\cite{wittig2012sabio} as of November 2023. 
This construction aims to provide an expanded and standardized resource for the field of enzyme kinetic parameter prediction, thereby promoting the systematic development and fair benchmarking of enzyme kinetic prediction models.

The datasets focus on three primary enzyme kinetic parameters: the turnover number ($k_\text{cat}$), Michaelis constant ($K_\text{m}$), and inhibition constant ($K_\text{i}$), based on in vitro measurements of wild-type enzymes. Specifically, the datasets include:
\begin{itemize}[leftmargin=2em]
    \item 23,151 entries for Kinetic Parameter $k_\text{cat}$;
    \item 41,174 entries for Kinetic Parameter $K_\text{m}$;
    \item 11,929 entries for Kinetic Parameter $K_\text{i}$.
\end{itemize}

The construction of these datasets introduces several improvements:

\paragraph{\ding{182} Data Selection and Standardization:}
When multiple experimental measurements exist for the same enzyme-substrate pair, the highest value is retained for $k_\text{cat}$ to capture optimal reaction conditions. For $K_\text{m}$ and $K_\text{i}$, the geometric mean is used, reflecting better consistency and stability for binding affinity measurements.

\paragraph{\ding{183} Diversity of Enzyme Sequences:}
These datasets significantly expand enzyme sequence diversity compared to previous datasets. They cover a broader range of enzyme families, particularly at the first Enzyme Commission (EC) level, without overrepresenting any specific categories. This is an important step towards fair and comprehensive benchmarking.

\paragraph{\ding{184} Link to Structural Data:}
Each entry is linked to a predicted three-dimensional structure of the corresponding enzyme. In cases where structures are not available in AlphaFold-2.0~\cite{Jumper2021AlphaFold}, ESMFold~\cite{Lin2023ESM2} is used to generate these models. This ensures that structural context is available for all entries, further enriching the dataset for model training.

\paragraph{\ding{185} Supporting Fair Benchmarking:}
By standardizing and expanding the sequence and structure coverage, the dataset offers a more representative resource for enzyme kinetic parameter prediction. This makes it well-suited for developing and fairly evaluating learning-based models in the field.

Overall, this expanded dataset not only enhances the diversity of enzyme sequences but also ensures consistency and stability in model training by using a representative value for each enzyme-substrate pair. Compared to prior works like DLKcat~\cite{Li2022DLKcat} and TurNup~\cite{Kroll2023TurNup}, our dataset covers a broader range of enzyme families and provides a more comprehensive foundation for large-scale enzyme kinetic predictions.

\subsection{Details OOD Dataset from~\cite{Shen2024EITLEM}}
To evaluate the out-of-distribution (OOD) generalization performance of our model, we utilize two enzyme kinetic datasets focused on $k_{\text{cat}}$ and $K_{\text{m}}$, compiled from the BRENDA and SABIO-RK databases as collected by~\cite{Shen2024EITLEM} in Table~\ref{tab:sup_dataset}. Each data entry contains enzyme sequences, substrate information, and experimentally measured kinetic parameters.

\paragraph{\ding{182} Data Curation and Filtering:}
The initial collections contained 143,473 entries for $k_{\text{cat}}$ and 75,149 entries for $K_{\text{m}}$. To ensure biological relevance, every protein sequence was mapped to a unique UniProt ID, and sequences lacking a valid or unambiguous ID were excluded. Substrate names were cross-referenced with KEGG~\cite{kanehisa2000kegg} and PubChem~\cite{wang2009pubchem} to obtain standardized 2D molecular structures. These substrates were then converted into canonical SMILES using RDKit~\cite{landrum2013rdkit}. Any substrates with ambiguous or unresolvable identities were discarded to maintain chemical consistency.

\paragraph{\ding{183} Inclusion of Mutant Enzymes:}
Mutant enzyme variants were retained in the dataset. While $K_{\text{m}}$ showed minimal differences between wild-type and mutant enzymes, $k_{\text{cat}}$ distributions revealed that mutant enzymes typically exhibit significantly reduced catalytic activity and efficiency compared to wild-type counterparts.

\paragraph{\ding{184} Final dataset composition:}
After rigorous screening, the final dataset comprised 35,147 $k_{\text{cat}}$ records and 29,321 $K_{\text{m}}$ records, with one-tenth allocated as a test set for evaluation.

\paragraph{\ding{185} 3D Structure Generation for OOD Testing:}
Notably, we use the $k_{\text{cat}}$ and $K_{\text{m}}$ test sets from~\cite{Shen2024EITLEM} for out-of-distribution generalization testing. The corresponding 3D structures for the enzymes are generated using OpenFold~\cite{ahdritz2024openfold} or ESMFold~\cite{Lin2023ESM2}, ensuring that the structural data is consistent with the latest advancements in protein structure prediction.

\begin{table*}[th!]
\centering
\resizebox{\textwidth}{!}{
\begin{tabular}{l|c|GGGG |HHHH |IIII}
\multirow{2}{*}{Index} & \multirow{2}{*}{Class Name} &
\multicolumn{4}{c|}{Kinetic Endpoint: $k_\text{cat}$} & \multicolumn{4}{c|}{Kinetic Endpoint: $K_\text{m}$} & \multicolumn{4}{c}{Kinetic Endpoint: $K_\text{i}$} \\
& & \cellcolor{white}R$^2$\(\uparrow\) & \cellcolor{white}PCC\(\uparrow\) & \cellcolor{white}RMSE\(\downarrow\) & \cellcolor{white}MAE\(\downarrow\)  & \cellcolor{white}R$^2$\(\uparrow\) & \cellcolor{white}PCC\(\uparrow\) & \cellcolor{white}RMSE\(\downarrow\) & \cellcolor{white}MAE\(\downarrow\) & \cellcolor{white}R$^2$\(\uparrow\) & \cellcolor{white}PCC\(\uparrow\) & \cellcolor{white}RMSE\(\downarrow\) & \cellcolor{white}MAE\(\downarrow\) \\
\thickhline
\multicolumn{14}{l}{\textcolor{gray}{{\textbf{(a)} \textit{\textbf{Ankh3-5.7B~\cite{Alsamkary2025Ankh3} \textit{w}/ ERBA}}}}}\\ 
EC-1 & Oxidoreductases & 0.39 & 0.68 & 1.25 & 0.89 & 0.51 & 0.78 & 0.83 & 0.61 & 0.53 & 0.73 & 1.29 & 0.96 \\
EC-2 & Transferases & 0.44 & 0.65 & 1.23 & 0.87 & 0.55 & 0.74 & 0.78 & 0.58 & 0.39 & 0.65 & 1.27 & 0.97 \\
EC-3 & Hydrolases & 0.45 & 0.70 & 1.13 & 0.85 & 0.53 & 0.73 & 0.85 & 0.60 & 0.29 & 0.60 & 1.72 & 1.27 \\
EC-4 & Lyases & 0.52 & 0.73 & 1.20 & 0.91 & 0.60 & 0.79 & 0.73 & 0.55 & 0.54 & 0.74 & 1.36 & 0.99 \\
EC-5 & Isomerases & 0.42 & 0.62 & 1.41 & 0.95 & 0.60 & 0.76 & 0.85 & 0.67 & 0.65 & 0.82 & 1.15 & 0.91 \\
EC-6 & Ligases & 0.50 & 0.71 & 1.01 & 0.79 & 0.30 & 0.55 & 1.12 & 0.83 & 0.27 & 0.51 & 1.38 & 1.08 \\
\hline\hline
\multicolumn{14}{l}{\textcolor{gray}{{\textbf{(b)} \textit{\textbf{ProtT5-3B~\cite{elnaggar2021prottrans} \textit{w}/ ERBA}}}}}\\ 
EC-1 & Oxidoreductases & 0.40 & 0.71 & 1.44 & 0.99 & 0.60 & 0.79 & 0.77 & 0.58  & 0.55 & 0.73 & 1.25 & 0.98 \\
EC-2 & Transferases & 0.48 & 0.73 & 1.25 & 0.90 & 0.51 & 0.69 & 0.79 & 0.58 & 0.42 & 0.69 & 1.39 & 1.03 \\
EC-3 & Hydrolases & 0.42 & 0.69 & 1.20 & 0.89 & 0.50 & 0.71 & 0.94 & 0.69 & 0.46 & 0.68 & 1.41 & 1.04 \\
EC-4 & Lyases & 0.48 & 0.70 & 1.21 & 0.85 & 0.53 & 0.71 & 0.87 & 0.65 & 0.69 & 0.83 & 1.07 & 0.77 \\
EC-5 & Isomerases & 0.46 & 0.70 & 1.29 & 0.90 & 0.63 & 0.80 & 0.80 & 0.61 & 0.65 & 0.81 & 1.10 & 0.89 \\
EC-6 & Ligases & 0.54 & 0.79 & 0.92 & 0.67 & 0.40 & 0.65 & 0.97 & 0.71 & 0.41 & 0.65 & 1.30 & 0.97 \\ \hline\hline
\multicolumn{14}{l}{\textcolor{gray}{{\textbf{(c)} \textit{\textbf{ESM2-650M~\cite{Lin2023ESM2} \textit{w}/ ERBA}}}}}\\ 
EC-1 & Oxidoreductases & 0.35 & 0.62 & 1.37 & 1.05 & 0.43 & 0.67 & 1.01 & 0.77 & 0.51 & 0.72 & 1.28 & 0.98 \\
EC-2 & Transferases & 0.39 & 0.66 & 1.34 & 1.02 & 0.34 & 0.62 & 0.96 & 0.74 & 0.35 & 0.60 & 1.39 & 1.05 \\
EC-3 & Hydrolases & 0.40 & 0.65 & 1.32 & 1.01 & 0.33 & 0.60 & 1.05 & 0.80 & 0.31 & 0.65 & 1.71 & 1.29 \\
EC-4 & Lyases & 0.44 & 0.69 & 1.28 & 0.98 & 0.31 & 0.58 & 1.00 & 0.77 & 0.46 & 0.74 & 1.47 & 1.10 \\
EC-5 & Isomerases & 0.43 & 0.65 & 1.32 & 0.99 & 0.53 & 0.77 & 0.90 & 0.81 & 0.71 & 0.87 & 1.12 & 0.88 \\
EC-6 & Ligases & 0.50 & 0.71 & 0.99 & 0.79 & 0.33 & 0.57 & 1.09 & 0.85 & 0.32 & 0.60 & 1.41 & 1.08 \\
\hline\hline
\multicolumn{14}{l}{\textcolor{gray}{{\textbf{(d)} \textit{\textbf{Ours (ESM2-3B~\cite{Lin2023ESM2} \textit{w}/ ERBA)}}}}}\\ 
EC-1 & \textit{Oxidoreductases} & 0.37 & 0.62 & 1.32 & 0.95  & 0.55 & 0.74 & 0.90 & 0.69  & 0.58 & 0.77 & {1.19} & {0.92} \\
EC-2 & \textit{Transferases} & 0.46 & 0.69 & 1.22 & {0.87} & 0.53 & 0.73 & {0.81} & {0.61}  & 0.38 & 0.65 & 1.37 & 0.99 \\
EC-3 & \textit{Hydrolases}& 0.43 & 0.68 & {1.18} & 0.88  & {0.51} & 0.72 & 0.90 & 0.65 & 0.48 & 0.71 & 1.50 & 1.10 \\
EC-4 & \textit{Lyases} & {0.50} & 0.71 & 1.27 & 0.92    & 0.56 & {0.75} & {0.81} & {0.61} & {0.65} & {0.81} & {1.19} & {0.84} \\
EC-5 & \textit{Isomerases}& 0.46 & {0.73} & 1.39 & 0.94    & 0.61 & {0.78} & {0.84} & {0.64}& {0.67} & {0.83} & {1.20} & 0.95 \\
EC-6 & \textit{Ligases} & {0.53} & {0.77} & {0.96} & {0.71} & {0.34} & 0.59 & 1.02 & 0.76 & 0.34 & 0.62 & 1.39 & 1.03 \\ \hline
\end{tabular}
}
\caption{Performance Comparison Across Enzyme Commission Classes. The table compares the performance of different backbone models and the impact of ERBA on kinetic parameter prediction for six EC classes: EC 1 (Oxidoreductases), EC 2 (Transferases), EC 3 (Hydrolases), EC 4 (Lyases), EC 5 (Isomerases), and EC 6 (Ligases). Metrics include R$^2$, PCC, RMSE, and MAE for $k_\text{cat}$, $K_\text{m}$, and $K_\text{i}$, showcasing the effect of model size and architecture on the predictive accuracy across enzyme classes.}
\label{tab:sup_ec}
\end{table*}

\section{Enzyme Commission Distribution}
This section explores the performance of various models across different Enzyme Commission (EC) classes, focusing on the influence of model architecture and the impact of our \textbf{ERBA (Enzyme-Reaction Bridging Adapter)} in improving predictions for enzyme kinetic parameters.

\subsection{Overview of EC Class Distributions}
Enzyme Commission (EC) classes group enzymes based on their catalytic functions, and understanding the distribution of these enzymes across various EC classes is crucial for predicting their kinetic parameters accurately. The datasets used in this experiment cover a wide range of EC classes, from EC 1 (Oxidoreductases) to EC 6 (Ligases), and include kinetic data for $k_\text{cat}$, $K_\text{m}$, and $K_\text{i}$.
Each class includes different numbers of entries, with EC 3 (Hydrolases) containing the most data, and EC 6 (Ligases) containing fewer. The purpose of this analysis is to evaluate how different model backbones and ERBA impact the prediction performance across these classes. It is notable that EC 7 (Translocases) is excluded from the analysis due to insufficient data.

\subsection{Effect of Different Model Architectures Across EC Classes}
Table~\ref{tab:sup_ec} presents a comparison of various models across six EC classes, with a focus on the performance metrics R$^2$, PCC, RMSE, and MAE for $k_\text{cat}$, $K_\text{m}$, and $K_\text{i}$. We observe that larger backbone models, such as Ankh3-5.7B~\cite{Alsamkary2025Ankh3} and ProtT5-3B~\cite{prot5}, outperform smaller models, like ESM2-650M, across all EC classes, suggesting that larger models are better equipped to handle the complexities of enzyme-substrate interactions. In EC 1 (Oxidoreductases), Ankh3-5.7B achieves a higher R$^2$ score for $k_\text{cat}$ (0.52) compared to ESM2-650M (R$^2$ = 0.45). This indicates that larger backbone models are better at capturing the intricate relationships between enzyme sequences and substrate properties.

\subsection{Impact of Model Size within the ESM Architecture}
Focusing specifically on the ESM2 architecture, we examine how different model sizes affect performance. As shown in (c) and (d), increasing the model size from ESM2-650M to ESM2-3B leads to noticeable improvements in performance across EC classes. For instance, in EC 5 (Isomerases), the R$^2$ score for $k_\text{cat}$ improves from 0.48 with ESM2-650M to 0.50 with ESM2-3B. This performance increase reflects the enhanced capacity of larger models to process more complex enzyme-substrate interactions, which are particularly important in enzyme families with diverse substrates like EC 5 and EC 6.

\subsection{Conclusion}
The results demonstrate that larger backbone models, such as Ankh3-5.7B and ProtT5-3B, perform better across diverse EC classes, highlighting the importance of model size in capturing complex enzymatic processes. In the case of ESM2, increasing the model size from ESM2-650M to ESM2-3B significantly enhances performance, especially in EC classes with diverse substrates, such as EC 5 (Isomerases). Furthermore, the introduction of ERBA provides a substantial boost to model performance by incorporating both substrate recognition and structural adaptation. This underscores the effectiveness of staged conditioning in improving enzyme kinetic predictions and enhancing the model’s generalization capabilities.

\begin{figure*}[t!]
\centering
\includegraphics[width=0.9\linewidth]{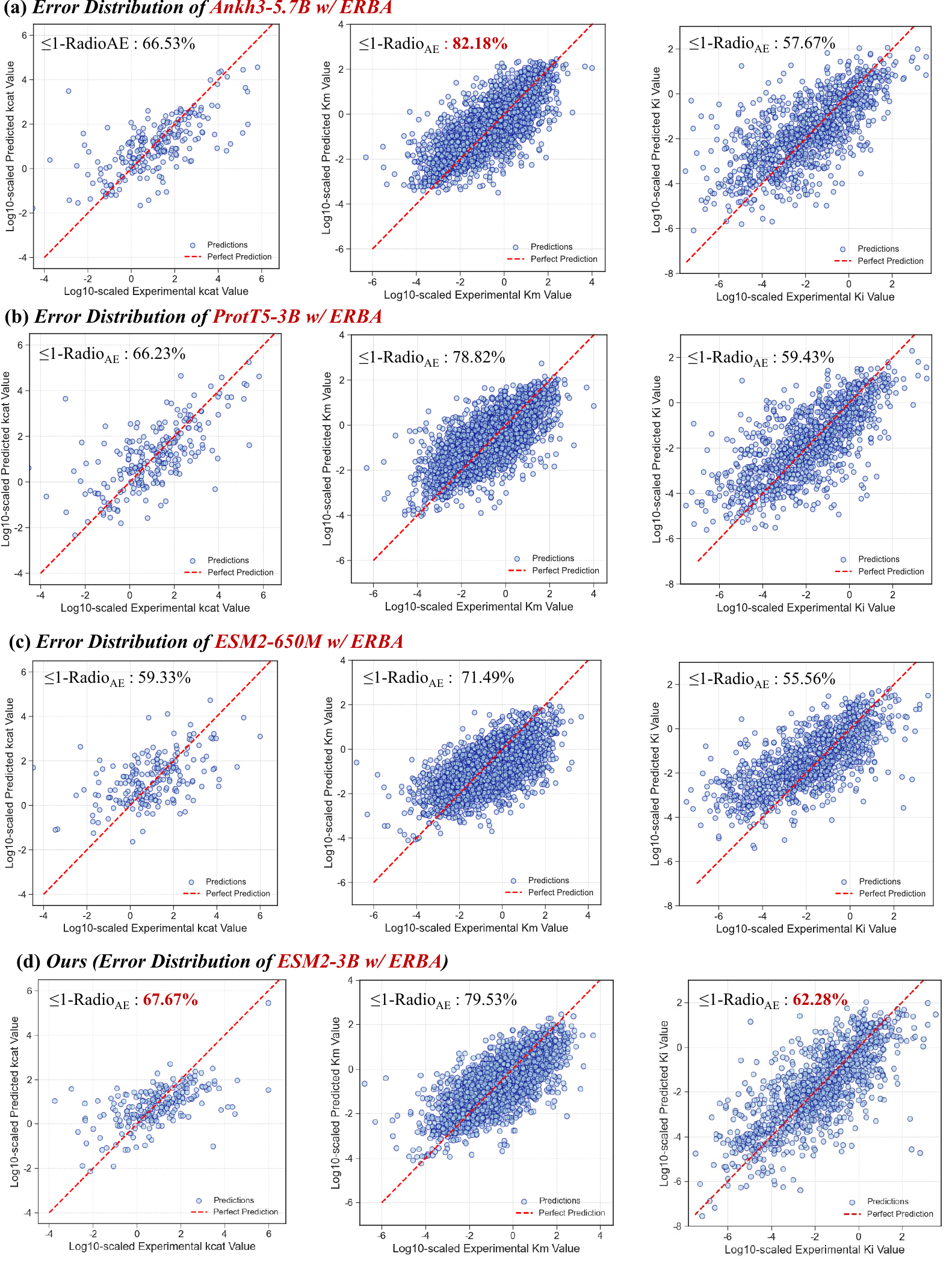}
\vspace{-1.5em}
\caption{
Error Distribution Comparison Across Different Backbone Models and ESM Sizes. It shows the error distribution of predicted versus experimental values for three kinetic parameters ($k_\text{cat}$, $K_\text{m}$, and $K_\text{i}$) across four backbone models: (a) Ankh3-5.7B~\cite{Alsamkary2025Ankh3}, (b) ProtT5-3B~\cite{elnaggar2021prottrans}, (c) ESM2-650M~\cite{Lin2023ESM2}, and (d) ESM2-3B~\cite{Lin2023ESM2}, each augmented with ERBA. The plots show the proportion of predictions with absolute error less than or equal to 1, denoted as 1-$\text{Radio}_{\text{AE}}$.
} 
\label{fig:error_dist}
\end{figure*}

\section{Error Distribution Analysis}
\subsection{Overview of Error Distributions}
The error distributions across different models (Ankh3-5.7B, ProtT5-3B, ESM2-650M, and ESM2-3B) and their corresponding performance with ERBA are analyzed in terms of absolute prediction error.
As shown in Figure~\ref{fig:error_dist}, all models show an overall improvement when enhanced with ERBA, particularly in terms of 1-$\text{Radio}_{\text{AE}}$ values, which represent the percentage of predictions with an absolute error smaller than 1.

\subsection{Performance of Ankh3-5.7B and Its Unique Behavior on $K_\text{m}$}
Notably, the Ankh3-5.7B model, while showing lower overall performance compared to ESM2-3B (as observed in the manuscript), performs significantly better in terms of $K_\text{m}$ predictions. Specifically, the Ankh3-5.7B model achieves a high 1-$\text{Radio}_{\text{AE}}$ value of 78.82\% for $K_\text{m}$ predictions, indicating a higher proportion of predictions with small errors despite having lower overall R$^2$ and PCC scores. This suggests that Ankh3-5.7B may be better at making precise predictions within a narrower range of $K_\text{m}$ values, while other models may struggle with broader generalization across the $K_\text{m}$ space.

\subsection{ESM2-3B: Best Overall Performance with ERBA}
The ESM2-3B model consistently outperforms all other backbones, particularly for $k_\text{cat}$ and $K_\text{m}$. The error distributions show that the ESM2-3B with ERBA achieves the highest 1-$\text{Radio}{\text{AE}}$ values (67.67\% for $k_\text{cat}$, 79.53\% for $K_\text{m}$, and 62.28\% for $K_\text{i}$), demonstrating its robust performance across all three kinetic endpoints. This confirms the effectiveness of the ESM2-3B backbone combined with ERBA for enzyme kinetic parameter prediction.

\subsection{Comparison with Smaller ESM2 Variants}
Interestingly, while the larger models like ESM2-3B excel overall, the smaller versions (such as ESM2-650M) show a reduction in the percentage of predictions with smaller errors, especially for $K_\text{m}$. This highlights the trade-off between model size and accuracy for out-of-distribution generalization. Larger models tend to perform better overall but may exhibit more variability in specific kinetic parameter predictions, particularly in smaller enzyme classes or in scenarios with less data.

\section{Explainable Analysis}
\begin{figure}[t!]
\begin{center}
\includegraphics[width=1\linewidth]{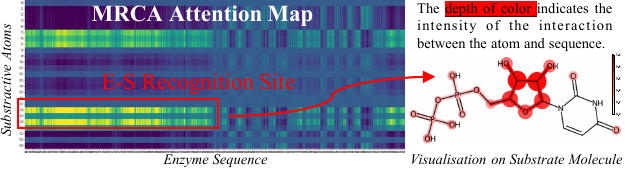}
\vspace{-2.2 em}
\caption{Visualization of MRCA.}
\label{fig:MRCA}
\vspace{-2.3em}
\end{center}
\end{figure}

\begin{figure}[t!]
\begin{center}
\includegraphics[width=1\linewidth]{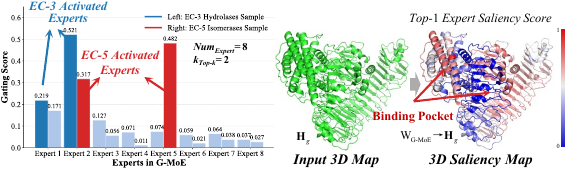}
\vspace{-2.2 em}
\caption{Visualization of G-MoE.}
\vspace{-2.3em}
\label{fig:G-MoE}
\end{center}
\end{figure}

We further analyze the interpretability of our method through the visualization of module behaviors.
We further analyze the interpretability of our method through the visualization of module behaviors.
In Figure~\ref{fig:MRCA}, MRCA  produces highly localized attention regions on the sequence and assigns the highest scores to chemically functional groups on the substrate. 
In Figure~\ref{fig:G-MoE}, the gating distribution is sparse and dynamically relies on enzyme characteristics, while 3D saliency map shows that top-1 expert responses cluster around the adaptive binding pocket.

\section{Additional Ablation Studies}

\begin{table}[t!]
\centering
\resizebox{\columnwidth}{!}{
\begin{tabular}{c|c|cccc }
 &\multirow{2}{*}{Method}&  \multicolumn{4}{c}{Kinetic Endpoint: $K_\text{i}$} \\
& & R$^2$↑ & PCC↑ & RMSE↓ & MAE↓\\
\thickhline
\multirow{2}{*}{Stage 1}&\textit{w}/ Concat \& Self-Atention~\cite{vaswani2017attention} &0.47 &0.69 &1.44 &1.10 \\
& \textbf{\textit{w}/ MRCA (Ours)} &\textbf{0.61} &\textbf{0.78} &\textbf{1.26} &\textbf{0.91} \\
\end{tabular}
}
\caption{Stage 1 (Recognition) ablation on $K_\text{i}$: Replacing \textit{Concat \& Self-Attention} with our \textbf{MRCA} markedly improves $R^2$ and PCC while reducing RMSE/MAE under identical settings.}
\label{tab:MRCA}
\end{table}
\begin{table}[t!]
\centering
\resizebox{\columnwidth}{!}{
\begin{tabular}{c|c|cccc }
 &\multirow{2}{*}{Method}&  \multicolumn{4}{c}{Kinetic Endpoint: $K_\text{i}$} \\
& & R$^2$↑ & PCC↑ & RMSE↓ & MAE↓\\
\thickhline
\multirow{2}{*}{Stage 2}&\textit{w}/ MoE~\cite{shazeer2017outrageously}  & 0.50 &0.71 &1.39 &1.05 \\
& \textbf{\textit{w}/ G-MoE (Ours)} &\textbf{0.61} &\textbf{0.78} &\textbf{1.26} &\textbf{0.91} \\
\end{tabular}
}
\caption{Stage 2 (Adaptation) ablation on $K_\text{i}$: \textit{standard MoE} \textit{vs.} our \textbf{G-MoE} that routes by pocket geometry; geometry-aware experts yield consistent gains across all metrics.}
\label{tab:MOE}
\end{table}

\begin{table}[t!]
\centering
\resizebox{\columnwidth}{!}{
\begin{tabular}{c|c|cccc }
 &\multirow{2}{*}{Method}&  \multicolumn{4}{c}{Kinetic Endpoint: $K_\text{i}$} \\
& & R$^2$↑ & PCC↑ & RMSE↓ & MAE↓\\
\thickhline
\multirow{2}{*}{Optimization}&\textit{w}/ $\mathcal{L}_2$ Loss~\cite{widrow1988l2loss} &0.48 &0.69 &1.42 &1.08 \\
& \textbf{\textit{w}/ ESDA (Ours)} &\textbf{0.61} &\textbf{0.78} &\textbf{1.26} &\textbf{0.91} \\
\end{tabular}
}
\caption{Objective ablation on $K_\text{i}$: \textit{plain $\mathcal{L}_2$} \textit{vs.} \textbf{ESDA}-regularized objective. Distribution alignment in RKHS stabilizes fine-tuning and improves R$^2$/PCC while lowering RMSE/MAE.
}
\label{tab:Loss}
\end{table}

To isolate the contribution of each component, we ablate on $K_\text{i}$ prediction under identical preprocessing, backbone, and training schedule. Across all studies, each module provides measurable gains and the full model is consistently best.

\subsection{Stage~1: Does MRCA Help Recognition?}
We replace the common \textit{Concat \& Self-Attention}~\cite{vaswani2017attention} fusion with our \textbf{MRCA}, as shown in Table~\ref{tab:MRCA}, which explicitly conditions enzyme tokens on substrate tokens in the PLM latent space before any structural input. MRCA improves $R^2$ from 0.47 to \textbf{0.61} (+0.14 / $\uparrow$29.8\%), PCC from 0.69 to \textbf{0.78} (+0.09 / $\uparrow$13.0\%), and reduces RMSE from 1.44 to \textbf{1.26} ($\downarrow$12.5\%) and MAE from 1.10 to \textbf{0.91} ($\downarrow$17.3\%). 
Results indicate that conditioning the PLM on substrate identity early yields a cleaner recognition signal than shallow concatenation, which underuses PLM priors.

\subsection{Stage~2: Is Geometry-aware MoE Necessary?}  
As shown in Table~\ref{tab:MOE}, we compare a \textit{standard MoE}~\cite{shazeer2017outrageously} against our \textbf{G-MoE}, which routes by pocket geometry and applies pocket-local, low-rank adaptation. G-MoE raises $R^2$ 0.50$\rightarrow$\textbf{0.61} (+0.11 / $\uparrow$22.0\%) and PCC 0.71$\rightarrow$\textbf{0.78} (+0.07 / $\uparrow$9.9\%), while lowering RMSE 1.39$\rightarrow$\textbf{1.26} ($\downarrow$9.4\%) and MAE 1.05$\rightarrow$\textbf{0.91} ($\downarrow$13.3\%).
This validates that the geometry-routed experts specialize to distinct conformational regimes (pocket size/shape/flexibility), outperforming a single shared adaptation rule.

\subsection{Does ESDA Stabilize Fine-tuning?}  
We replace the plain $\mathcal{L}_2$ regression with our \textbf{ESDA} objective, reported in Table~\ref{tab:Loss}, which aligns the distributions of $\mathbf{H}^{(0)}$ (sequence), $\mathbf{H}^{(1)}$ (seq+substrate), and $\mathbf{H}^{(2)}$ (seq+substrate+structure) to the PLM manifold via RKHS MMD. ESDA improves $R^2$ 0.48$\rightarrow$\textbf{0.61} (+0.13 / $\uparrow$27.1\%) and PCC 0.69$\rightarrow$\textbf{0.78} (+0.09 / $\uparrow$13.0\%), and reduces RMSE 1.42$\rightarrow$\textbf{1.26} ($\downarrow$11.3\%) and MAE 1.08$\rightarrow$\textbf{0.91} ($\downarrow$15.7\%). 
Thus, the distribution alignment curbs feature flooding from strong structural cues and preserves pretrained biochemical semantics during adaptation.

\subsection{Overall}
Gains accumulate across stages: MRCA supplies substrate-aware recognition, G-MoE injects pocket-specific adaptation, and ESDA keeps both anchored to the PLM’s semantic space, yielding the best accuracy and lowest errors on $K_\text{i}$.

\end{document}